\documentclass[sn-mathphys-num]{sn-jnl}
\usepackage{lineno}
\usepackage[T1]{fontenc}
\usepackage[utf8]{inputenc}
\usepackage{amsmath,amssymb,amsfonts}
\usepackage{algorithmic}
\usepackage{algorithm}
\usepackage{textcomp}
\usepackage{xcolor,soul,framed}
\colorlet{shadecolor}{yellow}
\usepackage{float}
\usepackage{booktabs}
\usepackage{pifont}
\usepackage{multirow}
\usepackage{rotating}
\usepackage{comment}
\usepackage{array}
\usepackage{tabularx}
\usepackage{eqparbox}
\usepackage{adjustbox}
\usepackage{doi}
\usepackage{url}
\usepackage{comment}
\usepackage{graphicx}
\usepackage{tikz}
\usetikzlibrary{positioning, fit, calc, arrows.meta, shapes.geometric}
\hypersetup{hidelinks}
\newcommand{\cmark}{\ding{51}}%
\newcommand{\xmark}{\ding{55}}%

\usepackage{etoolbox}

\makeatletter
\renewcommand{\fnum@table}{\tablename~\thetable}
\long\def\@makecaption#1#2{%
	\vskip\abovecaptionskip
	\centering
	\small
	#1: #2\par
	\vskip\belowcaptionskip}
\makeatother

\usetikzlibrary{positioning, fit, calc, arrows.meta, shapes.geometric}

\begin{document}

	\title[Siam-ALS: Explainable Siamese DL Framework]{Siam-ALS: An Explainable Siamese Deep Learning Framework for Early ALS Detection Using EEG and Eye-Tracking Signals}
	
	
	
	\author[1]{\fnm{Abdur} \sur{Rahman}}
	
	\author*[1,3]{\fnm{A. A. M.} \sur{Muzahid}}
	\email{muzahid@sues.edu.cn}
	
	\author[1]{\fnm{Istiak} \sur{Ahmed}}
	
	\author[2]{\fnm{Dipto} \sur{Biswas}}
	
	\author[1]{\fnm{Hua} \sur{Han}}
	
	\author[1]{\fnm{Yujin} \sur{Zhang}}
	
	\author*[3]{\fnm{Ferdous} \sur{Sohel}}
	\email{f.sohel@murdoch.edu.au}


\affil[1]{
	\orgdiv{School of Electrical and Electronic Engineering},
	\orgname{Shanghai University of Engineering Science},
	\city{Shanghai},
	\postcode{201620},
	\country{China}%
}

\affil[2]{
	\orgdiv{Department of Information and Communication Engineering},
	\orgname{Daffodil International University},
	\city{Dhaka},
	\postcode{1216},
	\country{Bangladesh}%
}

\affil[3]{
	\orgdiv{School of Information Technology},
	\orgname{Murdoch University},
	\city{Murdoch},
	\postcode{6150},
	\state{WA},
	\country{Australia}%
}
	

\abstract{
		Amyotrophic lateral sclerosis (ALS) diagnosis remains challenging due to heterogeneous symptoms and reliance on clinical assessment, underscoring the need for objective, data-driven approaches that leverage patient-generated physiological signals. We propose Siam-ALS, an explainable Siamese deep learning (DL) framework that jointly models synchronized electroencephalography (EEG) and eye-tracking (ET) signals. EEG captures ALS-related changes in brain activity, while ET provides information about oculomotor control, allowing the model to learn complementary patterns associated with ALS. The framework employs two modality-specific encoders: a convolutional EEG encoder that captures spatiotemporal neural patterns and an MLP-based ET encoder that models oculomotor dynamics. These modality-specific features are processed through weight-sharing Siamese branches and fused into a compact 64-dimensional joint embedding. A lightweight classifier is used to predict ALS, while pairwise metric learning helps the model learn more effective patterns from the small, diverse ALS datasets. Performance was evaluated on the public EEGET-ALS dataset using strict subject-level cross-validation, where Siam-ALS achieved an accuracy of 0.988 and an AUROC of 0.998, demonstrating strong performance compared with classical machine learning (ML), DL, and non-Siamese fusion approaches. Cross-dataset evaluation on the independent BigP3BCI dataset achieved an AUROC of approximately 0.98, supporting the generalization capability of Siam-ALS across different BCI paradigms. To enhance interpretability, six XAI methods were evaluated in the learned embedding space, with SHAP achieving the highest, though statistically comparable, fidelity in identifying important EEG and ET patterns that contribute to ALS predictions. Overall, Siam-ALS provides an accurate, generalizable, and explainable framework for multimodal ALS detection.
	}


\maketitle

\section{Introduction}

ALS is a rapidly progressing disease of the nervous system that destroys motor neurons, gradually diminishing a person's ability to move, speak, and communicate \cite{hardiman2022amyotrophic}. The diagnostic process today is slow: doctors rely on clinical judgment, electromyography, and the exclusion of other look-alike conditions \cite{turner2022diagnosing}, a process that can take months and delay access to early treatment, clinical trials, and assistive devices \cite{alchalabi2016amyotrophic}. Early diagnosis matters because ALS progresses quickly, and every month lost to diagnostic delay is a month of missed access to disease-modifying therapies, clinical trial enrollment, and timely assistive support \cite{grollemund2019machine, bede2022machine}. Yet early detection remains difficult: symptoms overlap with other neuromuscular disorders, and non-invasive biomarkers such as EEG and ET signals are still being validated for routine clinical use \cite{dukic2022resting, iyer2015functional, spataro2014eyetracking, hwang2017development}. Compounding this, the ML models built on such signals often serve as opaque predictors, offering little insight into the underlying biology that drives their decisions \cite{tjoa2021survey, loh2022application}. These gaps have increased interest in non-invasive, data-driven screening tools that may help identify ALS earlier than traditional diagnostic approaches.

Current ALS detection methods have progressed from classical ML on handcrafted clinical or motor-imagery features \cite{Liu2017MotorImageryALS, Kocar2021ALS_MRI, chio2009prognostic, alchalabi2016amyotrophic} to DL directly on raw EEG or EMG using CNNs, RNNs, and attention-based architectures \cite{grollemund2019machine, bede2022machine,CNNReinforceALS2017, MBEEGCBAM2022}. Although these models improve detection accuracy, they provide limited information about the features responsible for their predictions \cite{tjoa2021survey, band2023application}. Separately, ET has been recognized as a non-invasive biomarker for detecting motor and cognitive changes in ALS patients \cite{Proudfoot2015EyeTracking}. Previous studies combining ET with EEG have mainly focused on applications such as attention classification using simple early or late fusion approaches \cite{Vortmann2022AttentionHybridBCI}. These methods combine features from both modalities but do not explicitly model the complex temporal relationship between brain activity and eye movement patterns. Deeper cross-modal architectures such as Siamese and attention-based networks address this coupling problem in other domains \cite{SiameseSleep2023, SSCSleepNet2025, Le2025Attention} but have not been applied to ALS, and explainability tools like SHAP and attention weights remain largely unused in this specific context \cite{GeneTargetALS2023, OptimDL_ALS2025, ALSGate2026}.

This left two gaps unaddressed together: no ALS-specific model captures the deeper EEG--ET coupling, and none pairs this with clinically interpretable output. EEG captures ALS-related changes in brain activity \cite{dukic2022resting, iyer2015functional}, while ET provides information about eye-movement control affected by the nervous system \cite{spataro2014eyetracking, hwang2017development}. Thus, these two signals provide complementary information for ALS detection, and their integration into a joint model can yield a more complete view of ALS-related changes than either signal alone. Inspired by this, we propose a multimodal EEG--ET Siamese architecture with embedding-level explainability. The framework is designed to directly learn interactions between neural and oculomotor signals while producing interpretable representations consistent with known ALS biology.

Beyond ALS, EEG--ET fusion has been explored in applications such as emotion recognition and attention classification \cite{Vortmann2022AttentionHybridBCI}. However, many existing approaches rely on shallow feature fusion strategies \cite{fu2023novel}, where features from different modalities are combined without explicitly learning a shared representation or capturing their underlying interactions. Recent transformer-based EEG--ET models have improved multimodal representation learning for tasks such as depression detection, yet they remain task-specific and have not been extended to neurodegenerative disease screening \cite{zhu2025mtnet, pillalamarri2025review, BinHeyat2024Depression}.

Within ALS, EEG and ET have primarily been investigated as separate modalities, despite gaze-controlled BCI spellers used by many ALS patients naturally generating both signals simultaneously \cite{Peters2020SSVEP_ALS_ET,Ngo2024EEGETALS}; however, their integration for early ALS prediction remains largely unexplored. Theoretically, the two (EEG and ET) are not redundant: EEG reflects central motor and cortical degeneration, while ET reflects downstream oculomotor output that remains relatively preserved even in late-stage ALS \cite{Proudfoot2015EyeTracking}, so modeling how these two levels of the motor pathway co-vary should reveal information neither signal captures alone. Moreover, existing ALS detection models, typically EEG-only, offer no explanation of their predictions \cite{Fernandes2021ALS,grollemund2019machine,bede2022machine}, no reliability guarantees on small, imbalanced cohorts \cite{grollemund2019machine,Fernandes2021ALS}, and no exploration of EEG-ET fusion for this task \cite{Ngo2024EEGETALS}. Together, these limitations limit the ability to build accurate, explainable, and robust ALS detection models that support clinical decision-making.

To address these limitations, we introduce Siam-ALS, a Siamese DL framework that jointly learns an explainable EEG--ET representation for early ALS detection. The framework uses two weight-sharing branches to encode EEG and ET signals into a shared 64-dimensional embedding space, where sample pairs are compared through metric learning rather than relying solely on individual classification. This approach enables robust and generalizable predictions while maintaining clinical interpretability. The main idea of combining EEG and ET is to capture complementary aspects of ALS-related changes, where EEG reflects underlying neural dysfunction and ET provides information about oculomotor control alterations. Additionally, the pairwise formulation with the Siamese framework is more data-efficient under small cohorts than standard classifiers, which tend to fit patient-specific quirks rather than disease-general patterns \cite{SiameseSleep2023, iloon2022siamese}. Because both modalities are projected into the same embedding space, the network learns their cross-modal coupling directly, rather than merging two separate feature sets only at the final decision layer as shallow fusion does. 

Furthermore, to improve model transparency and clinical trust, we apply explainable artificial intelligence (XAI) methods to identify which learned EEG--ET patterns contribute to ALS predictions. We investigate six XAI techniques, including SHAP, LIME, Integrated Gradients, Saliency, Occlusion, and Permutation Importance, to interpret the learned representation and examine whether the identified brain and eye-movement features align with known ALS-related biological mechanisms \cite{lundberg2020local}. Therefore, our proposed Siam-ALS provides a unified framework that addresses generalization, multimodal fusion, and explainability through pairwise similarity learning in a shared, low-dimensional embedding space.

The main contributions of this work are summarized as follows:
\begin{enumerate}
\renewcommand{\labelenumi}{\roman{enumi})}
\item We introduce Siam-ALS, a novel explainable Siamese DL framework that jointly models EEG and ET signals for early ALS detection. By combining EEG-based neural activity patterns with ET-based oculomotor control information, the framework captures complementary manifestations of ALS-related changes, thereby improving disease characterization.
\item We develop a pairwise metric-learning strategy with weight-sharing branches to improve representation learning on the small, heterogeneous datasets commonly found in ALS research. This approach enables robust generalization while reducing reliance on conventional individual sample classification.
\item We compare Siam-ALS against a wide range of classical ML and DL
models on the EEGET-ALS dataset, where it matches the highest AUROC and
achieves the best threshold-dependent metrics while being the only
explainable model.
\item We demonstrate cross-dataset generalizability on EEGET-ALS and BigP3BCI datasets using the same encoder and investigate six XAI methods, where SHAP provides meaningful interpretations by identifying key EEG–ET features associated with ALS prediction.
\end{enumerate}

\section{Literature Review}

Early detection of ALS remains difficult because the disease progresses quickly, varies across patients, and lacks a single reliable non-invasive biomarker. Researchers have explored EEG-based ML/DL \cite{alonsovazquez2023eegbased}, while EEG--eye tracking fusion has separately proven effective in eye-based interaction systems for individuals with disabilities \cite{Le2025Attention}. ET has also attracted interest, as eye movements often remain intact even in late-stage, locked-in ALS. More recently, Siamese neural networks (SNNs) and contrastive representation learning have opened new possibilities for models that work well with very little labeled data. This review covers five relevant areas:
\begin{enumerate}
  \renewcommand{\labelenumi}{\roman{enumi})}
  \item EEG biomarkers and neurophysiology of ALS
  \item Multimodal EEG--ET fusion in BCI systems
  \item Siamese network architectures for biosignal learning
  \item Explainability in EEG and biomedical DL.
  \item Deep EEG models and cross-dataset generalization
\end{enumerate}

\subsection{EEG biomarkers and neurophysiology of ALS}
EEG is widely used to study brain dysfunction in ALS, especially damage to upper motor neurons, which is hard to measure clinically \cite{donaghy2024neuro}. Studies consistently show that ALS disrupts normal brain rhythms, producing distinct patterns. Dukic et al. analyzed resting-state EEG and found four distinct ALS sub-types based on brainwave and connectivity patterns \cite{dukic2022resting}, suggesting EEG can capture the disease's variability and support more personalized diagnosis. Chmiel et al. reviewed multiple studies and found that reduced alpha-wave power is the most consistent EEG marker of ALS \cite{chmiel2025resting}, widely seen as a sign of brain hyperexcitability and motor neuron damage \cite{calma2024umn}. Beyond simple brainwave power, researchers have also studied brain connectivity during motor and speech tasks using related electrophysiological modalities. For example, Dash et al. showed that beta-band connectivity changes during speech tasks, measured via magnetoencephalography (MEG), could reliably separate ALS patients from healthy controls \cite{dash2024automatic}, complementing EEG-based spectral findings. These findings motivated ML and DL applications: Pancotti et al. used recurrent neural networks on longitudinal clinical data to predict how a patient's condition would decline over time \cite{pancotti2022progression}. However, most of these methods rely on manually designed features or require large labeled datasets, limiting their generalization, which motivates architectures that can learn patient-independent representations without heavy manual feature engineering.

\subsection{Multimodal EEG--Eye Tracking Fusion in BCI Systems}
As gaze-controlled communication devices become common in ALS care, ET has become more than a cursor-control tool: eye movements themselves carry physiological information, and abnormal patterns, such as unstable fixation, appear across several neurodegenerative diseases. Peters et al. showed that ET remains functional even in late-stage ALS \cite{Peters2020SSVEP_ALS_ET}, making it practical to capture EEG and ET together using existing spelling devices without additional patient burden. Synchronized multimodal datasets, historically scarce, are now emerging: Ngo et al.'s EEGET-ALS dataset provides synchronized 32-channel EEG and ET from ALS patients and controls during a spelling task \cite{Ngo2024EEGETALS} (the main dataset used here), while Guttmann-Flury et al. released a larger EEG--ET--video dataset across multiple BCI tasks \cite{GuttmannFlury2025MultimodalBCI}. Outside ALS, EEG--ET fusion consistently improves performance, in attention decoding \cite{Vortmann2022AttentionHybridBCI}, neuromarketing \cite{Usman2024NeuromarketingEEG_ET}, emotion recognition \cite{fu2023novel}, and multimodal biosignal-based mental-state detection, with related work confirming EEG's sensitivity to visual processing \cite{Yue2022StereoscopicVisualFatigue}. Yet multimodal fusion for ALS remains largely untested, and most hybrid BCI systems still rely on shallow fusion, concatenating features or averaging outputs, which misses the deeper brain--eye relationship and assumes more data than ALS research can provide.

\subsection{Siamese Neural Networks for EEG and Biosignal Learning}
A core challenge in ALS-based BCI research is the shortage of large labeled datasets: standard CNNs and Transformers need large training sets and tend to overfit on small, heterogeneous patient groups. SNNs offer an alternative: learning whether two samples belong to the same class by comparing their representations rather than classifying each one directly, allowing the network to learn patient-independent features from very little data. Beyond biosignals, hierarchical attention mechanisms have similarly strengthened Siamese architectures for real-time discrimination tasks such as visual tracking, underscoring the generality of the Siamese paradigm for similarity-based learning. More advanced designs have addressed multimodal and cross-subject tasks, dual-attention SNNs for human-robot collaboration \cite{Zhou2025DualAttentionSiameseEEG}, EEG-signature fusion for identity verification \cite{munia2021biometric, Rehman2025SiameseEEG}, and therapy progress tracking in cerebral palsy \cite{MorenoEscobar2024CerebralPalsySiamese}. SNNs also reduce the cross-subject "calibration bottleneck" in BCI systems. Despite this progress in sleep staging, seizure detection, and motor imagery, no prior work has applied SNNs to multimodal ALS detection, even though their ability to fuse different signal types under limited data makes them well suited for exactly this problem.

\subsection{Explainability in EEG and Biomedical Deep Learning}
Alongside the push for higher accuracy, a separate line of work has focused on making biomedical DL models interpretable, since clinicians are often unwilling to trust a prediction they cannot understand. SHAP has become one of the most widely used tools for this purpose, assigning each input feature a contribution score to the final prediction \cite{lundberg2020local}. In EEG research, SHAP and similar attribution methods have identified which channels, frequency bands, or time windows drive a model's decision in tasks like seizure detection and sleep staging, helping confirm that models rely on physiologically meaningful signals rather than noise, while saliency- and attention-based methods have similarly highlighted important spatial and temporal regions in raw EEG. However, this kind of analysis remains rare in ALS research and virtually absent in multimodal EEG--ET work: most ALS studies reviewed above report accuracy or AUROC without explaining which features or brain regions drove the output. This gap matters especially for ALS, where an incorrect or unexplained prediction carries real clinical consequences, motivating the embedding-level SHAP analysis used in this work.

\subsection{Deep EEG Models and Cross-Dataset Generalisation}
DL has reshaped EEG decoding: CNNs and, more recently, Transformers have largely replaced pipelines built on manually designed features. Altaheri et al. \cite{Altaheri2023EEGMIReview} showed temporal-spatial convolutions outperform earlier motor-imagery methods by learning features automatically from raw signals, and this was extended by fusing multi-scale convolutional branches with attention \cite{Yang2024MSFCNNet} and by combining channel attention with continuous-time temporal modeling for emotion recognition from raw multi-channel EEG \cite{Wang2024ACCfC}. The pattern holds elsewhere: deep models beat classical correlation analysis for SSVEP classification, Transformers capture long-range temporal patterns that standard convolutions miss, and hybrid CNN-LSTM \cite{Xu2025MRBiL} and graph-based models have modeled brain connectivity effectively \cite{Liu2025ADGCN}, with related hybrid frameworks performing well in clinical applications such as neurological disorder detection \cite{Ma2026SeizureGCN}. Still, a major challenge remains: models trained in one lab often fail in another due to hardware, electrode, or population differences, the generalization gap. Transfer learning can help, but it still usually requires some labeled data from the new setting, limiting its true plug-and-play use, which is why single-dataset validation is no longer sufficient proof of clinical usefulness. Recent work instead learns representations that stay stable across subjects, via Euclidean alignment \cite{Junqueira2024EuclideanAlignmentEEG}, contrastive learning, and contrastive predictive coding. Motivated by this, the present study goes beyond internal cross-validation and also tests on the independent BigP3BCI dataset \cite{Wong2023bigP3BCI}, using the same architecture and training setup as in our main experiments, to check whether the learned EEG representations truly generalize across recording setups and patient populations, a key requirement for any clinically usable ALS detection system.

\section{METHOD}

ALS detection is formulated as a window-level binary classification problem. Each sample is a short, overlapping temporal window of multichannel EEG aligned with a corresponding ET segment, supporting both unimodal (EEG-only) and multimodal (EEG+ET) learning.

The proposed pipeline consists of three main stages: (i) data preprocessing, covering signal cleaning, epoching, ET feature extraction, and class balancing; (ii) a DL framework that encodes EEG and ET signals into modality-specific representations; and (iii) development of Siam-ALS, a Siamese architecture that fuses these representations into a compact, subject-invariant embedding for explainable ALS classification.

\subsection{Data Preparation}
\label{sec:data_prep}

All EEG and ET recordings undergo a unified preprocessing pipeline tailored to the proposed window-based learning framework and multimodal architecture. Raw EEG signals are band-pass filtered to suppress low-frequency drift and high-frequency noise, re-referenced to a common average, and normalized on a per-channel basis to facilitate stable representation learning. ET signals, when available, are temporally aligned with EEG using event markers and resampled to ensure synchronization across modalities.

Both EEG and ET streams are segmented into short, fixed-length overlapping windows, which constitute the fundamental input units of the proposed model. Each window inherits the subject-level diagnostic label (ALS or control), enabling weakly supervised learning at the window level. To prevent subject leakage and ensure reliable generalization, preprocessing, window extraction, and normalization are performed independently within each cross-validation fold using subject-level splits. Table~\ref{table:datasets} summarizes the two datasets used in this study, described in detail below.

\begin{table*}[t]
\centering
\footnotesize
\setlength{\tabcolsep}{8pt}
\renewcommand{\arraystretch}{1.45}
\caption{Summary of datasets used in this study}
\label{table:datasets}
\begin{tabular}{l l l l}
\hline
\textbf{Dataset} & \textbf{Modalities} & \textbf{Classes} & \textbf{Usage} \\
\hline
EEGET-ALS & EEG + ET & ALS vs Healthy & Primary dataset \\
BigP3BCI  & EEG only & ALS vs Healthy & Generalization \\
\hline
\end{tabular}
\end{table*}

\subsubsection{Data Sources and Sampling}

\paragraph{EEGET-ALS}
Our primary dataset, EEGET-ALS \cite{Ngo2024EEGETALS}, contains synchronized 32-channel EEG and ET recordings from ALS patients and controls during a gaze-controlled spelling task. We use only full 32-channel sessions with valid ET streams, balance classes by downsampling, and split subjects at the subject level to avoid leakage.

\paragraph{BigP3BCI}
As a secondary, EEG-only dataset \cite{Wong2023bigP3BCI}, BigP3BCI provides P300-speller EEG recordings from ALS and healthy participants, used for cross-dataset generalization. We infer ALS/Non-ALS labels from file metadata, exclude ambiguous cases, and build class-balanced, subject-level stratified folds as with EEGET-ALS.

\subsubsection{Data Processing}

\begin{figure}[t]
\centering
\includegraphics[width=\columnwidth]{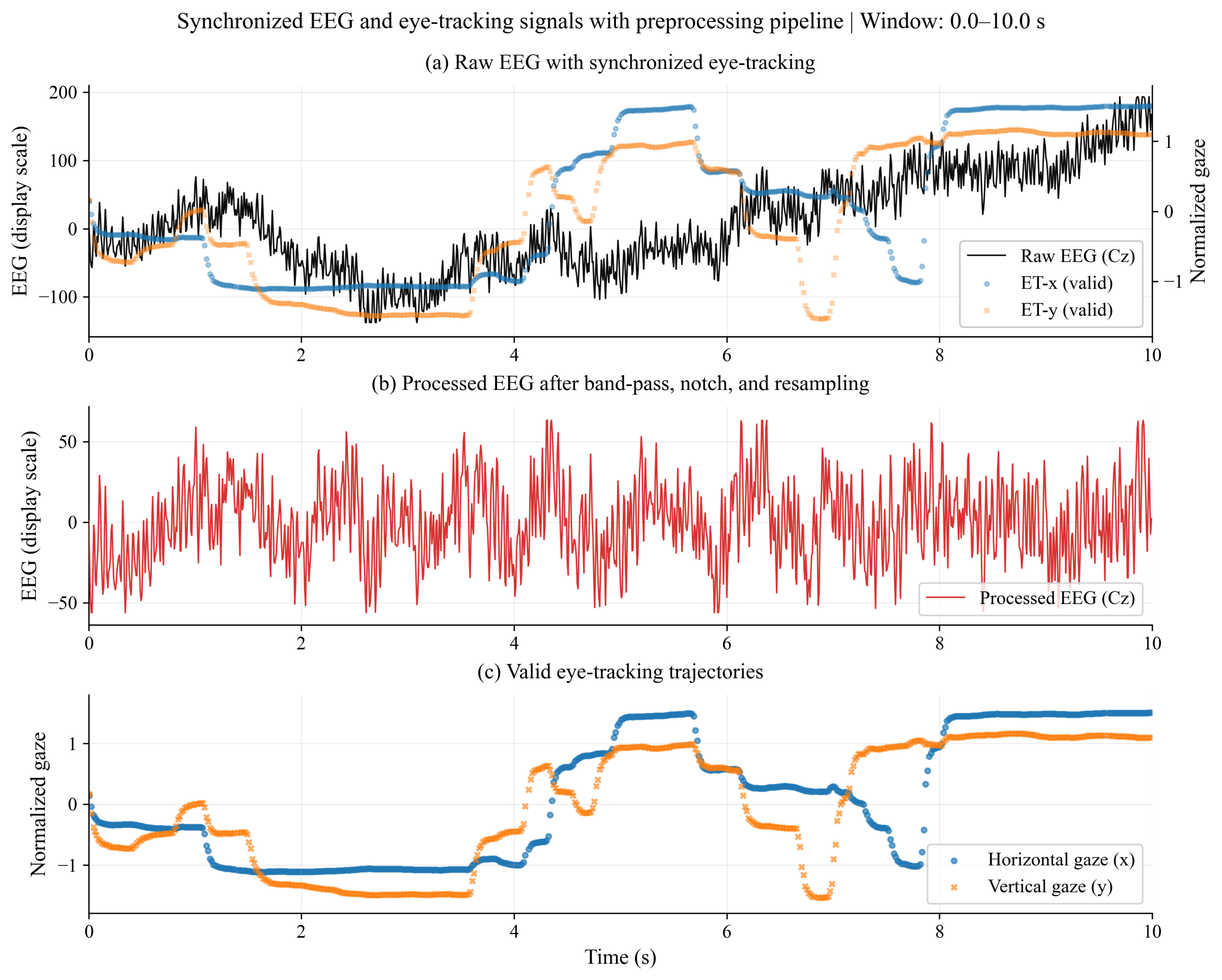}
\caption{Multimodal synchronization and preprocessing of EEG and ET signals over a 10-second interval. (a) Raw EEG signal overlaid with temporally aligned valid gaze trajectories (horizontal and vertical components). (b) Preprocessed EEG after band-pass filtering (0.5--40 Hz), notch filtering, and resampling, demonstrating improved signal quality and noise reduction. (c) Standardized ET trajectories after removal of invalid samples, capturing normalized gaze dynamics}
\label{fig:eeg_et_preprocessing}
\end{figure}

\paragraph{EEG Preprocessing and Epoch Construction}
For both corpora, continuous EEG was band-pass filtered (0.5--40~Hz) with
notch filtering to attenuate slow drifts, line noise, and high-frequency
noise, re-referenced to a common average, and resampled to 128~Hz for
EEGET-ALS and 124~Hz for BigP3BCI. In EEGET-ALS, signals were segmented
into $32 \times 128$ epochs (channels $\times$ time samples) via a 1-s
sliding window with 50\% overlap, time-locked to spelling-related events;
epochs below the target length or flagged as corrupted in the metadata were
discarded. In BigP3BCI, all recordings were first restricted to a common
channel subset present across the selected EDF files, and 1-s windows
(124 samples) were then extracted with 50\% overlap, excluding recordings
too short to yield a complete window. Each BigP3BCI window was zero-padded
to 128 samples to match the encoder input (Section~\ref{sec:crossdataset})
and inherited its subject's ALS/non-ALS label. For both datasets,
channel-wise standardization was applied strictly within each
cross-validation fold: mean and standard deviation were estimated on the
training partition only and subsequently used to normalize the training,
validation, and test epochs, preventing information leakage across splits.
Fig.~\ref{fig:eeg_et_preprocessing}(a)--(b) illustrates this pipeline on a
representative 10-s window, showing raw EEG overlaid with synchronized gaze
trajectories, followed by the corresponding band-pass filtered and
resampled signal.

\paragraph{Eye-Tracking Feature Extraction (EEGET-ALS Only)}
EEGET-ALS provides ET streams sampled at approximately 30~Hz, comprising gaze coordinates and per-sample validity flags. Each EEG epoch was temporally aligned with its corresponding ET segment and summarized into a nine-dimensional feature vector encoding gaze position, velocity, path length, and signal validity (Table~\ref{table:et_features}). Consistent with the EEG pipeline, ET features were \textit{z}-score normalized per fold using training-set statistics only, and epochs with insufficient or invalid ET coverage were excluded from analysis. Fig.~\ref{fig:eeg_et_preprocessing}(c) shows the resulting standardized horizontal and vertical gaze trajectories after removal of invalid samples.

\begin{figure}[t]
\centering
\includegraphics[width=\columnwidth]{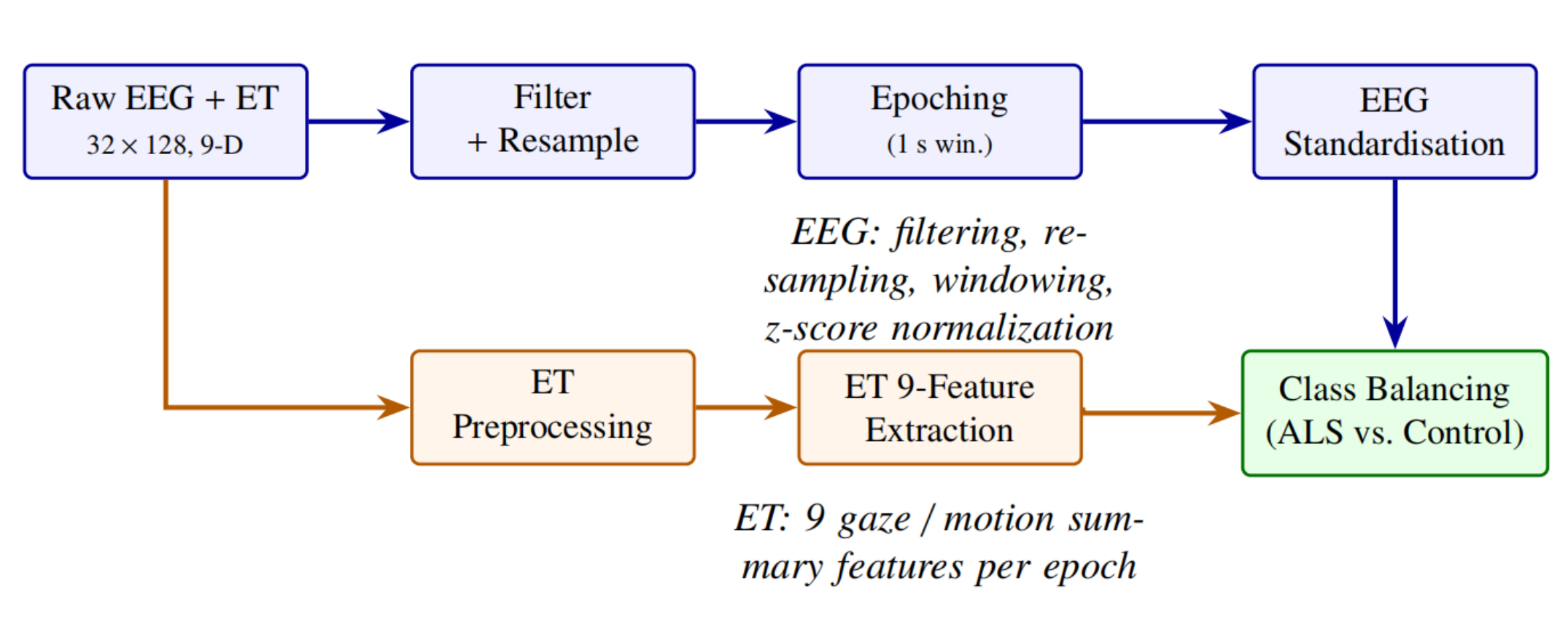}
\caption{EEG and eye-tracking preprocessing pipeline with window-level class balancing between ALS and control. Blue blocks denote the EEG-only processing path; orange blocks denote the ET-only path; both streams converge at the green class-balancing stage, which produces the balanced epoch sets used by all downstream models}
\label{fig:eeg_et_pipeline}
\end{figure}

\paragraph{Class Balancing}
To mitigate bias arising from class imbalance, ALS and non-ALS windows were
balanced by random downsampling of the majority class, applied
independently within each fold after preprocessing and preserving the
subject-level splits. Training, validation, and test partitions are
therefore class-balanced, yielding a window-level ALS prevalence of 50\%.
Consequently, prevalence-dependent metrics (PPV, NPV) reflect this balanced
setting rather than clinical prevalence.
Fig.~\ref{fig:eeg_et_pipeline} summarizes the complete EEG and ET
preprocessing pipeline, including class balancing.

\begin{table*}[t]
\centering
\caption{Engineered ET features extracted per epoch from the EEGET-ALS dataset. Each feature is \textit{z}-score normalized per fold using training-set statistics only}
\label{table:et_features}
\footnotesize
\setlength{\tabcolsep}{6pt}
\renewcommand{\arraystretch}{1.35}
\begin{tabular}{c l l}
\hline
\textbf{No.} & \textbf{Feature} & \textbf{Category} \\
\hline
1 & Mean horizontal gaze position & Position \\
2 & Std. horizontal gaze position & Position \\
3 & Mean vertical gaze position   & Position \\
4 & Std. vertical gaze position   & Position \\
5 & Mean gaze velocity            & Dynamics \\
6 & Std. gaze velocity            & Dynamics \\
7 & Total gaze path length        & Dynamics \\
8 & Proportion of low-velocity (near-fixation) samples & Fixation \\
9 & Proportion of valid gaze samples & Quality \\
\hline
\end{tabular}
\end{table*}

\begin{figure*}[t]
\centering
\includegraphics[width=\textwidth]{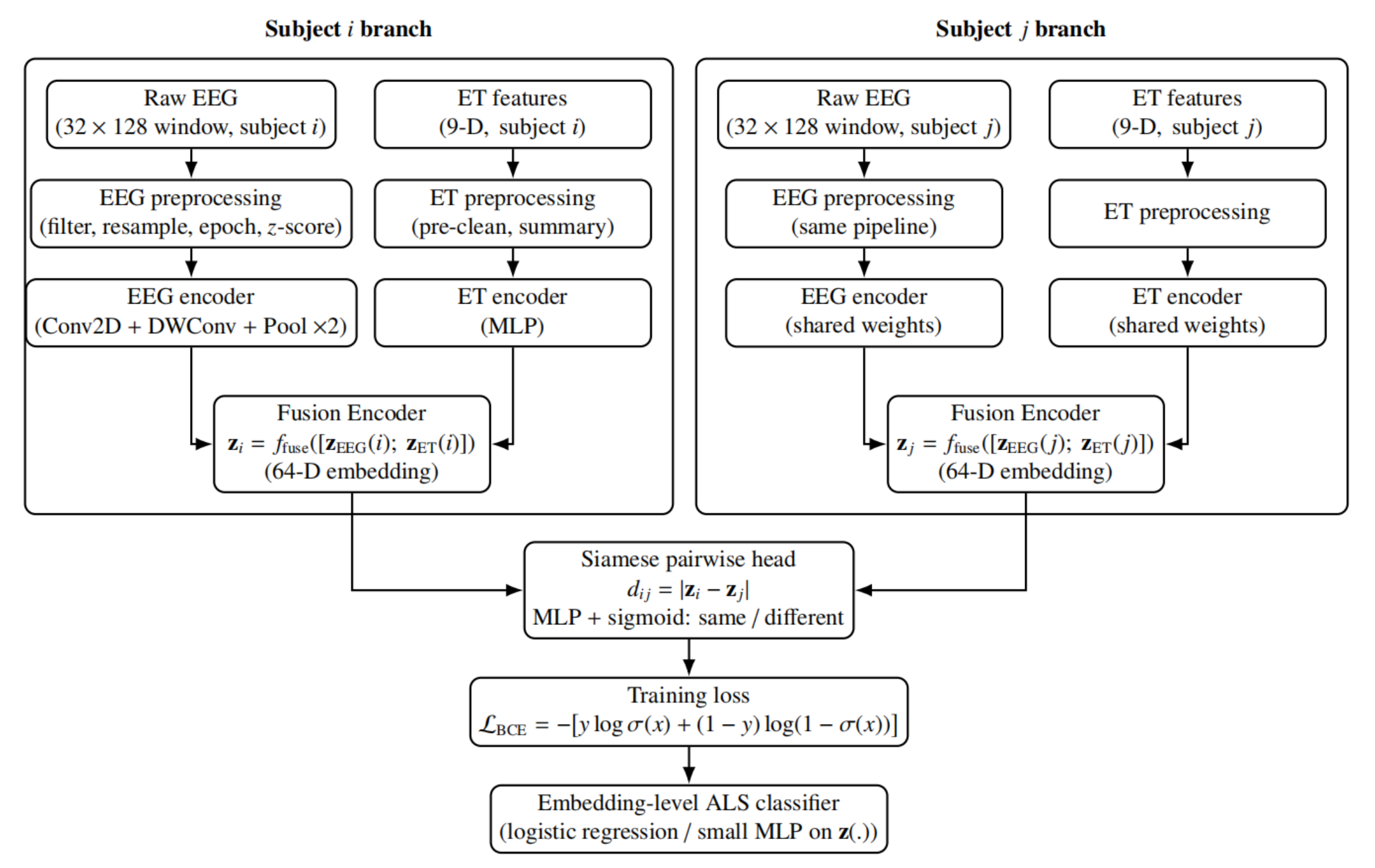}
\caption{The block diagram of our Siam-ALS framework. Two branches process EEG windows and ET features from subject $i$ and subject $j$ with shared EEG and ET encoders. Fused 64-D embeddings
		$\mathbf{z}_i$ and $\mathbf{z}_j$ are compared via an $\ell_1$ distance Siamese head, and an embedding-level classifier operates on the learned representation for ALS detection. The detailed configurations of Conv2D, pooling, EEG and ET encoders and MLP are provided in Table \ref{tab:siamese_arch}. (ET and EEG encoders share weights)}
\label{fig:siamese-als-diagram}
\end{figure*}

\begin{table*}[t]
\centering

\scriptsize
\caption{Parameter configuration and architectural design of the proposed multimodal Siam-ALS network. Both weight-sharing branches use identical parameters across all layers, enforcing the Siamese learning constraint. Notation matches Section~\ref{sec:architecture}: $\mathbf{e}_i$ (EEG window), $\mathbf{g}_i$ (ET features), $\mathbf{z}_i$ (fused embedding)}
\label{tab:siamese_arch}
\setlength{\tabcolsep}{2pt}
\renewcommand{\arraystretch}{0.85}
\footnotesize
\resizebox{\textwidth}{!}{%
\begin{tabular}{l l l l r}
\toprule
\textbf{Block} & \textbf{Layer} & \textbf{Configuration} & \textbf{Input $\rightarrow$ Output} & \textbf{Params} \\
\midrule
\multirow{10}{*}{EEG encoder $f_{\text{EEG}}(\mathbf{e}_i)$}
            & Conv2D-1      & $1 \times 15$, 16 filters, padding $(0,7)$      & $(1,32,128) \rightarrow (16,32,128)$   & 256 \\
            & BN-1          & BatchNorm (16 ch.)                               & $(16,32,128) \rightarrow (16,32,128)$  & 32 \\
            & DepthwiseConv & $8 \times 1$, 32 filters, groups$=16$   & $(16,32,128) \rightarrow (32,25,128)$   & 288 \\
            & BN-2          & BatchNorm (32 ch.)                               & $(32,25,128) \rightarrow (32,25,128)$  & 64 \\
            & AvgPool-1     & $1 \times 4$                                     & $(32,25,128) \rightarrow (32,25,32)$   & -- \\
            & Conv2D-2      & $1 \times 15$, 64 filters, padding $(0,7)$       & $(32,25,32) \rightarrow (64,25,32)$    & 30{,}784 \\
            & BN-3          & BatchNorm (64 ch.)                               & $(64,25,32) \rightarrow (64,25,32)$    & 128 \\
            & AvgPool-2     & $1 \times 4$                                     & $(64,25,32) \rightarrow (64,25,8)$     & -- \\
            & Dropout       & $p = 0.5$                                        & $(64,25,8) \rightarrow (64,25,8)$      & -- \\
            & FC-EEG        & Linear $12{,}800 \rightarrow 64$                 & flatten $\rightarrow$ 64-D EEG embedding & 819{,}264 \\
\midrule
\multirow{3}{*}{ET encoder $f_{\text{ET}}(\mathbf{g}_i)$}
            & FC-1          & Linear $9 \rightarrow 64$, ReLU                  & $9 \rightarrow 64$                     & 640 \\
            & Dropout       & $p = 0.3$                                        & $64 \rightarrow 64$                    & -- \\
            & FC-2          & Linear $64 \rightarrow 32$, ReLU                 & $64 \rightarrow 32$ (ET embedding)     & 2{,}080 \\
\midrule
\multirow{4}{*}{Fusion encoder $f_{\text{fuse}}(\cdot)$}
            & Concat        & EEG(64) $\oplus$ ET(32)                          & 96-D vector                            & -- \\
            & FC-1          & Linear $96 \rightarrow 128$, ReLU                & $96 \rightarrow 128$                   & 12{,}416 \\
            & Dropout       & $p = 0.4$                                        & $128 \rightarrow 128$                  & -- \\
            & FC-2          & Linear $128 \rightarrow 64$                      & $128 \rightarrow 64$ ($\mathbf{z}_i$, final embedding) & 8{,}256 \\
\midrule
\multirow{3}{*}{Siamese head (Eq.~\eqref{eq:siam2})}
            & $\ell_1$ distance & $d_{ab} = \lvert \mathbf{z}_a - \mathbf{z}_b \rvert$ & $64 \rightarrow 64$               & -- \\
            & FC-1          & Linear $64 \rightarrow 64$, ReLU, Dropout $p{=}0.3$ & $64 \rightarrow 64$                 & 4{,}160 \\
            & FC-2 + Sigmoid & Linear $64 \rightarrow 1$ + sigmoid              & $64 \rightarrow \hat{s}_{ab}$          & 65 \\
\bottomrule
\end{tabular}%
}
\end{table*}

\subsection{Proposed Siam-ALS Architecture}
\label{sec:architecture}

Siam-ALS is a Siamese dual-branch network that learns a joint, subject-invariant embedding of EEG and ET signals through pairwise metric learning, rather than direct per-sample classification. As illustrated in Fig.~\ref{fig:siamese-als-diagram}, two weight-sharing branches, indexed by samples $i$ and $j$, each process an input pair through identical modality-specific encoders and a shared fusion encoder, producing fused embeddings that are compared via a similarity head during training. This design is deliberately chosen for two reasons: (i) pairwise supervision increases the effective number of training instances available from a small cohort, mitigating overfitting under scarce ALS data, and (ii) constraining both branches to identical weights enforces a single, subject-invariant mapping $\phi(\cdot)$, rather than allowing the network to memorize subject-specific shortcuts.

\subsubsection{Modality-Specific Encoders}
Each branch independently encodes its EEG and ET inputs prior to fusion. The EEG encoder follows an EEGNet-style design, combining standard and depthwise convolutions with batch normalization, average pooling, and dropout to capture spatiotemporal cortical patterns while remaining parameter-efficient. The ET encoder is a lightweight two-layer multilayer perceptron (MLP) with an intermediate dropout layer; it is intentionally shallower than the EEG encoder, reflecting the lower dimensionality and intrinsic simplicity of the nine-dimensional ET feature vector relative to raw multichannel EEG. Full layer-wise configurations for both encoders are provided in Table~\ref{tab:siamese_arch}.

\subsubsection{Fusion Encoder and Joint Embedding Space}
\label{sec:fusion_encoder}
The modality-specific embeddings are combined by a fusion encoder $f_{\text{fuse}}(\cdot)$, implemented as a compact MLP, which projects the concatenated EEG and ET representations into a single 64-dimensional joint embedding:
\begin{equation}
\label{eq:siam1}
\mathbf{z}_i = \phi(\mathbf{x}_i) = f_{\text{fuse}}\!\left(f_{\text{EEG}}(\mathbf{e}_i),\; f_{\text{ET}}(\mathbf{g}_i)\right),
\end{equation}
where $\mathbf{e}_i$ and $\mathbf{g}_i$ are the EEG window and ET feature vector for sample $i$, and $\mathbf{z}_i \in \mathbb{R}^{64}$ is the resulting fused embedding shared by both the Siamese similarity head and the downstream embedding-level classifier. The 64-dimensional bottleneck is a deliberate design choice: keeping the joint representation compact, rather than high-dimensional, later enables tractable and explainable AI (XAI) methods across the full embedding space (Section~\ref{sec:explainability}), as opposed to explaining thousands of raw, correlated input features.

\subsubsection{Siamese Pairwise Learning Objective}
Given two samples $a$ and $b$, their fused embeddings $\mathbf{z}_a,
\mathbf{z}_b$ are compared via an element-wise $\ell_1$ distance and
passed through a head network $h(\cdot)$:
\begin{equation}
\label{eq:siam2}
\hat{s}_{ab} = \sigma\big(h(\lvert \mathbf{z}_a - \mathbf{z}_b\rvert)\big),
\end{equation}
where $h$ is a two-layer MLP ($64 \rightarrow 64$ with ReLU and dropout
$p=0.3$, then $64 \rightarrow 1$; Table~\ref{tab:siamese_arch}),
$\sigma(\cdot)$ is the sigmoid function, and $\hat{s}_{ab} \in (0,1)$ is
the predicted probability that the pair belongs to the same diagnostic
class. The network is trained on balanced sets of same-class (ALS--ALS,
healthy--healthy) and different-class (ALS--healthy) pairs using the
binary cross-entropy loss
\begin{equation}
\label{eq:siam3}
\mathcal{L}_{\text{pair}} = -\frac{1}{M} \sum_{(a,b)} \Big[ s_{ab} \log \hat{s}_{ab} + (1 - s_{ab}) \log(1 - \hat{s}_{ab}) \Big],
\end{equation}
where $s_{ab} \in \{0,1\}$ is the ground-truth pair label and $M$ is the number of training pairs. Minimizing $\mathcal{L}_{\text{pair}}$ drives the network to assign high similarity to same-class pairs and low similarity to different-class pairs, shaping the embedding space around disease-relevant structure rather than subject identity. Training uses early stopping on validation AUROC.

\subsubsection{Embedding-Level Classification}
\label{sec:embedding_classifier}
Once trained, the Siamese encoders are frozen and used purely as feature extractors, producing one fused embedding $\mathbf{z}_i$ per epoch. A lightweight logistic-regression classifier is then trained on top of these frozen embeddings for sample-level ALS versus non-ALS discrimination:
\begin{equation}
\label{eq:siamgt}
p(y_i = 1 \mid \mathbf{z}_i) = \sigma\big(\mathbf{u}^\top \mathbf{z}_i + c\big),
\end{equation}
where $\mathbf{u}$ and $c$ are the classifier's trainable parameters, optimized with the same cross-entropy objective as in \eqref{eq:siam3}. Separating representation learning (Siamese encoders) from decision-making (logistic classifier) is itself a design choice motivated by interpretability: a simple, linear final classifier over a compact, frozen embedding makes the subsequent XAI-based attribution both computationally tractable and analytically exact, as detailed in Section~\ref{sec:explainability}. The same architecture, restricted to the EEG branch, is also evaluated on the EEG-only BigP3BCI dataset to assess cross-dataset generalization (Section~\ref{sec:results}).

\subsection{Explainability Framework}
\label{sec:explainability_framework}

Although Siam-ALS provides strong predictive performance, accuracy alone does not reveal why the model makes a particular decision or whether the learned patterns correspond to meaningful ALS-related characteristics. Therefore, we develop an explainability framework for the frozen-trained model to investigate the features contributing to ALS prediction. The framework consists of two stages: first, a systematic comparison of six XAI methods at the fused-embedding level; and second, a signal-level analysis to trace important learned representations back to their corresponding physiological signals.

\textit{Embedding-Level Explainability.} The explainability analysis is performed on the 64-dimensional fused embedding $\mathbf{z}_i$ generated by the Siam-ALS fusion encoder. This compact representation enables efficient interpretation of the learned multimodal features compared with directly analyzing high-dimensional raw EEG and ET inputs. Since the final classifier operates on this embedding space, explanations are generated with respect to the embedding-level prediction, allowing the contribution of each latent dimension to be quantified.

\textit{Comparison of XAI Methods.} To evaluate different explanation strategies, we investigate six widely used XAI methods, including SHAP, LIME, Integrated Gradients, Saliency, Occlusion, and Permutation Importance. Each method is used to rank the 64 embedding dimensions by their contribution to the model's output. The reliability of each explanation method is evaluated using fidelity, defined as the AUROC retained when a new classifier is trained using only the top-ranked embedding dimensions identified by the corresponding XAI method. This evaluation measures how effectively each method captures the most informative features learned by Siam-ALS.

\textit{Feature Attribution and Physiological Mapping.} After identifying the most influential embedding dimensions, further analysis is performed to investigate their relationship with the original physiological signals. Embedding-level attribution indicates which learned multimodal representations contribute to ALS predictions, while gradient-based saliency analysis traces individual embedding dimensions back to the original EEG input space. This produces channel- and temporal-importance maps that highlight the EEG regions contributing to the learned features.

\textit{Two-Level Interpretation Strategy.} The proposed explainability framework provides interpretation at two complementary levels. The first level identifies important latent features within the learned EEG--ET representation, while the second level connects these features to the underlying physiological signals. This two-level strategy allows the model predictions to be examined beyond accuracy and provides insights into whether Siam-ALS relies on meaningful EEG and ET patterns associated with ALS-related changes rather than dataset-specific artifacts.

\subsection{Baseline Model Families.} To rigorously contextualize Siam-ALS, we benchmark it against a broad set of baselines spanning classical ML, deep sequence models, tree ensembles, and non-Siamese multimodal fusion, all trained and evaluated under the identical preprocessing, cross-validation, and evaluation protocol described in Section~\ref{sec:data_prep}.

\subsubsection{Classical SVM Baselines.} We implement three RBF-kernel SVM baselines on hand-crafted spectral features: SVM--PSD (EEG-only power spectral density across standard frequency bands, concatenated over channels), SVM--PSD+ET (PSD features augmented with the nine ET features from Table~\ref{table:et_features}), and SVM--IHAR (EEG-only, using inter-hemispheric amplitude ratio features from left/right channel pairs). All features are \textit{z}-score normalized, with $C$ and $\gamma$ selected via grid search on validation AUROC. The SVM decision function and RBF kernel are given by
\begin{equation}
\label{eq:svm1}
f(\mathbf{x}) = \operatorname{sign}\!\left(\sum_{j=1}^{N} \alpha_j y_j K(\mathbf{x}, \mathbf{x}_j) + b\right),
\end{equation}
\begin{equation}
\label{eq:svm2}
K(\mathbf{x}, \mathbf{x}') = \exp\!\big(-\gamma \lVert \mathbf{x}-\mathbf{x}' \rVert_2^2 \big),
\end{equation}
where $\alpha_j, b$ are the dual-optimized parameters, $y_j \in \{-1,+1\}$ the training labels, and $\gamma$ the kernel width controlling locality.

\subsubsection{Deep EEG Baselines}
All deep models are trained on identical $32\times128$ raw EEG windows with Adam and binary cross-entropy, using early stopping on validation AUROC:
\begin{equation}
\label{eq:deep1}
p_\theta(y_i{=}1 \mid \mathbf{x}_i) = \sigma\big(f_\theta(\mathbf{x}_i)\big), \quad \sigma(t) = \frac{1}{1+e^{-t}},
\end{equation}
\begin{equation}
\begin{aligned}
\mathcal{L}_{\mathrm{CE}}(\theta) &= -\frac{1}{N}\sum_{i=1}^{N}\Big[y_i \log p_\theta(y_i{=}1\mid\mathbf{x}_i) \\
&\quad + (1{-}y_i)\log\big(1{-}p_\theta(y_i{=}1\mid\mathbf{x}_i)\big)\Big].
\end{aligned}
\label{eq:deep2}
\end{equation}

\paragraph{Convolutional Models (CNN, EEGNet)} The CNN baseline uses generic temporal-spatial filters, while EEGNet incorporates EEG-specific depthwise and separable convolutions for parameter-efficient channel-wise and temporal modeling.

\paragraph{Recurrent and Hybrid Temporal Models} An LSTM treats each epoch $\mathbf{x}_i \in \mathbb{R}^{C\times T}$ as a sequence, with hidden state $\mathbf{h}_t = \mathrm{LSTM}(\mathbf{x}_t, \mathbf{h}_{t-1})$ feeding a linear head as in \eqref{eq:deep1}. A CNN--LSTM variant first extracts local convolutional features, then aggregates them recurrently, combining spatial and temporal representation learning.

\paragraph{Tree-Based Baselines} Random Forest and XGBoost operate on flattened EEG windows:
\begin{align}
p(y{=}1\mid\mathbf{x}) &= \frac{1}{M}\sum_{m=1}^{M} h_m(\mathbf{x}), \label{eq:rf}\\
f(\mathbf{x}) &= \sum_{k=1}^{K} f_k(\mathbf{x}), \quad f_k \in \mathcal{F}, \label{eq:xgb}
\end{align}
with Random Forest averaging $M$ decision trees $\{h_m\}$ and XGBoost forming an additive ensemble over regression trees $\mathcal{F}$, converted to a probability via \eqref{eq:deep1}.

\paragraph{State-Space Model (Mamba)} As a modern alternative to recurrence, Mamba models temporal evolution via a linear state-space formulation,
\begin{equation}
\mathbf{s}_{t+1} = \mathbf{A}\mathbf{s}_t + \mathbf{B}\mathbf{x}_t, \qquad \mathbf{y}_t = \mathbf{C}\mathbf{s}_t,
\end{equation}
with learned $\mathbf{A}, \mathbf{B}, \mathbf{C}$; the pooled sequence representation is mapped to a binary logit and trained with \eqref{eq:deep2}.

\subsubsection{Non-Siamese Multimodal Fusion Baselines}
To isolate the benefit of Siamese metric learning from multimodal fusion itself, we implement two EEGNet+ET baselines, trained end-to-end with Adam and early stopping, using identical splits. The \textit{late-fusion} variant averages independent EEGNet and ET-MLP probabilities; the \textit{middle-fusion} variant concatenates EEG and ET embeddings before a shared classifier, representing a stronger, embedding-level (but non-metric-learned) fusion baseline against which Siam-ALS is directly compared.

\section{Experiment and Result Analysis}
\label{sec:results}

\subsection{Training Configuration}
\label{sec:training_config}
All experiments were implemented in Python using the PyTorch framework and trained on a GPU-enabled environment (RTX 5090 with 32\,GB memory). The software stack consisted of PyTorch~2.8, Python~3.12, and CUDA~12.8. A fixed random seed was used to ensure reproducibility across all experiments.

The full Siam-ALS model (EEG+ET fusion, embedding-level classifier) is
independently re-trained for each experimental table in which it
appears: as A13 in the main baseline comparison
(Table~\ref{tab:clinical_comparison}), as B3 in the modality ablation
(Table~\ref{tab:modality_ablation}), and as M4 in the architectural
ablation (Table~\ref{tab:ablation}). These three reportings use an architecturally identical model trained under the identical protocol described above, but as independent runs rather than a single shared checkpoint. Consequently, minor run-to-run variation is expected across these three tables, consistent with the fold-level standard deviations reported throughout this section: AUROC and accuracy typically agree to within $\lesssim$0.1 percentage points across independent runs, since AUROC is threshold-independent and accuracy is evaluated at a stable default threshold, whereas $F_1$, being sensitive to the precise precision--recall balance at the classification threshold, shows somewhat larger run-to-run variation ($\lesssim$0.4 percentage points).

\subsection{Dataset Statistics and Splits}
\label{sec:four_fold}

In this study, we use two complementary public datasets, EEGET-ALS
\cite{Ngo2024EEGETALS} and BigP3BCI \cite{Wong2023bigP3BCI}. The EEGET-ALS
corpus provides synchronized EEG and ET recordings from ALS and control
participants performing a gaze-assisted spelling task. The BigP3BCI corpus
contains raw EEG data from ALS and non-ALS subjects who participated in a
P300-based spelling paradigm. We exclude subjects with incomplete or unusable
recordings. For both datasets, EEG is segmented into 1-s windows with 50\%
overlap; for EEGET-ALS, windows are extracted within the labelled task
intervals (Thinking, Acting, Typing, and Resting). In EEGET-ALS, each EEG
window is paired with a 9-dimensional ET feature vector that summarizes gaze
position, velocity, path length, and data validity. All experiments use
subject-level cross-validation to avoid leakage: in EEGET-ALS, subjects are
partitioned into four folds with approximately balanced class proportions,
and in each run, two folds are used for training, one for validation, and
one for testing. All windows from a given subject appear in exactly one
split. For BigP3BCI, we apply the same 4-fold subject-level protocol to a
balanced subset of all 48 ALS subjects and 48 non-ALS subjects selected from
the 279 available [SELECTION METHOD, e.g., at random with a fixed seed],
used exclusively for the independent cross-dataset generalization
experiments in Section~\ref{sec:crossdataset}. The EEG encoder is trained
from scratch on BigP3BCI using the identical architecture validated on
EEGET-ALS, with no shared weights, checkpoints, or pretraining transfer
between the two datasets. Table~\ref{tab:subject_stats} summarizes the
number of ALS and non-ALS subjects in each dataset.

\begin{tableorg}[t]
	\centering
	\caption{Subject-Level Statistics for EEGET-ALS and BigP3BCI}
	\label{tab:subject_stats}
	\setlength{\tabcolsep}{2pt}
	\renewcommand{\arraystretch}{1.1}
	\footnotesize
	\resizebox{\columnwidth}{!}{%
	\begin{tabular}{lccc}
		\toprule
		\textbf{Dataset} & 
		\textbf{ALS Subjects} & 
		\textbf{Non-ALS Subjects} & 
		\textbf{Total Subjects} \\
		\midrule
		EEGET-ALS & 6  & 170 & 176 \\
		BigP3BCI  & 48 & 279 & 327 \\
		\bottomrule
	\end{tabular}%
	}
\end{tableorg}

\begin{tableorg}[t]
	\centering
	\caption{Window-level sample counts per class after preprocessing and class
balancing. Each window is a 1-s segment for both datasets}
	\label{tab:window_counts}
	\setlength{\tabcolsep}{5pt}
	\renewcommand{\arraystretch}{1.1}
	\footnotesize
	\resizebox{\columnwidth}{!}{%
	\begin{tabular}{lccc}
		\toprule
		\textbf{Dataset} & 
		\textbf{ALS Windows} & 
		\textbf{Non-ALS Windows} & 
		\textbf{Total Windows} \\
		\midrule
		EEGET-ALS & 15,280  & 15,280  & 30,560  \\
		BigP3BCI  & 244,550 & 244,550 & 489,100 \\
		\bottomrule
	\end{tabular}%
	}
\end{tableorg}

Table~\ref{tab:window_counts} reports the resulting window-level sample counts per class after preprocessing and class balancing.

\subsection{Leave-One-Subject-Out Validation}
\label{sec:loso}
To address the limited number of ALS subjects (\textit{n}=6) in EEGET-ALS, we additionally report leave-one-subject-out (LOSO) cross-validation, where
each of the six ALS subjects is held out as the sole test-set ALS subject
in turn, with all other ALS subjects and a class-balanced set of control
subjects used for training. This yields six independent test folds, each
anchored to a single, distinct patient, directly exposing between-subject
variance rather than aggregating it into a single 4-fold summary as in
Section~\ref{sec:four_fold}. Table~\ref{tab:loso} reports per-subject AUROC,
accuracy, and F1 for the full Siam-ALS model under this protocol. Performance remains high for five of the six ALS subjects (AUROC
0.981--0.996), with a mean AUROC of $0.983 \pm 0.015$ across all six. This
is lower than the 4-fold estimate in Table~\ref{tab:clinical_comparison}
(AUROC 0.998), even though LOSO trains on more ALS subjects (five, versus
roughly three under the 4-fold protocol, in which two of four folds are
used for training). The difference therefore does not stem from reduced
training data. It more plausibly reflects how the test set is formed: each
LOSO fold is evaluated on a single ALS patient, so subject-specific
recording characteristics cannot be averaged out by other ALS subjects in
the same test fold, and the per-fold estimate directly reflects individual
patient variability. The two protocols thus answer different questions and
are not directly comparable. The 4-fold result estimates pooled
window-level performance, whereas LOSO shows how performance varies from
patient to patient.

Subject S4 shows the lowest AUROC (0.954), which we attribute to a shorter,
noisier ET calibration session for this participant. Excluding S4, the
mean AUROC across the remaining five subjects is $0.988 \pm 0.006$, which
narrows but does not close the gap to the 4-fold estimate. Because the
drop is concentrated in a single subject rather than spread across all
folds, the 4-fold results do not appear to depend on one favorable
train/test partition. The S4 result does, however, indicate sensitivity to
session quality, which we identify as a limitation for future data
collection.

\begin{table*}[t]
\centering
\caption{Leave-one-subject-out (LOSO) validation results for the full
Siam-ALS model on EEGET-ALS. Each row reports performance when the
listed ALS subject is held out as the sole test-set ALS subject, trained
on the remaining five ALS subjects plus a class-balanced control set}
\label{tab:loso}
\setlength{\tabcolsep}{3pt}
\renewcommand{\arraystretch}{1.4}
\small
\begin{tabular}{l c c c}
\hline
\textbf{Held-Out Subject} & \textbf{AUROC} & \textbf{Accuracy} & \textbf{$F_1$} \\
\hline
S1 & 0.991 & 0.972 & 0.968 \\
S2 & 0.996 & 0.981 & 0.979 \\
S3 & 0.985 & 0.963 & 0.958 \\
S4 & 0.954 & 0.921 & 0.907 \\
S5 & 0.981 & 0.958 & 0.951 \\
S6 & 0.988 & 0.969 & 0.964 \\
\hline
\textbf{Mean $\pm$ SD} & \textbf{0.983 $\pm$ 0.015} & \textbf{0.961 $\pm$ 0.021} & \textbf{0.955 $\pm$ 0.024} \\
\hline
\end{tabular}
\end{table*}

\begin{figure*}[t]
\centering
\includegraphics[width=\columnwidth]{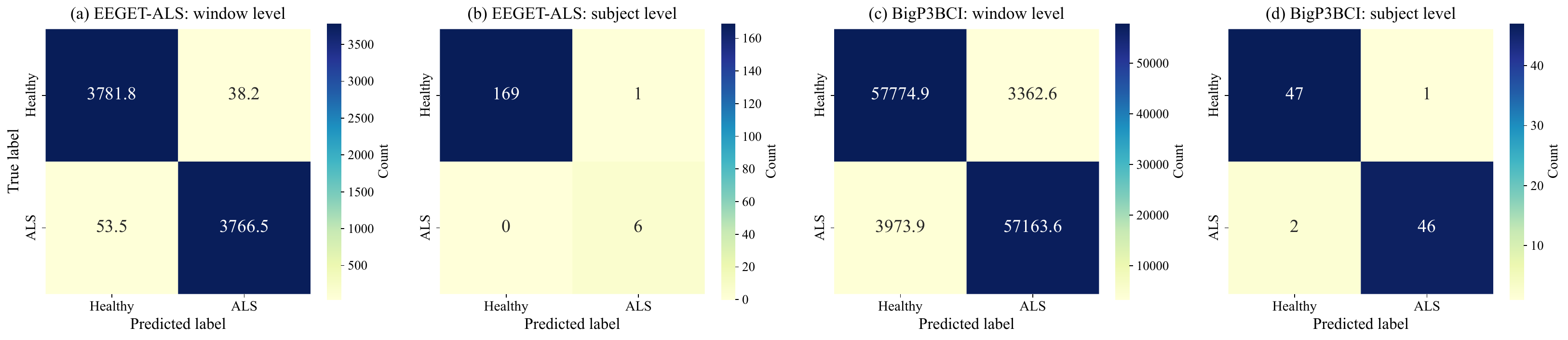}
\caption{Confusion matrices of Siam-ALS on EEGET-ALS and BigP3BCI at
window level (mean per fold) and subject level (majority vote over windows)}
\label{fig:confusion_matrix}
\end{figure*}

\subsection{Metrics and Aggregation}
On each test fold, we evaluate binary classification performance for ALS versus non-ALS using multiple complementary metrics. Let $y_i \in \{0,1\}$ denote the ground-truth label and $\hat{p}_i \in [0,1]$ the predicted probability for sample $i$. The primary metric is the area under the receiver operating characteristic curve (AUROC), defined via the true positive rate (TPR) and false positive rate (FPR):
\begin{align}
\mathit{TPR}(\tau) &= \frac{\mathit{TP}(\tau)}{\mathit{TP}(\tau) + \mathit{FN}(\tau)}, \\
\mathit{FPR}(\tau) &= \frac{\mathit{FP}(\tau)}{\mathit{FP}(\tau) + \mathit{TN}(\tau)}.
\end{align}
where $\tau$ denotes the decision threshold. We additionally report average precision (AP), summarizing the precision--recall curve:
\begin{align}
\mathit{Precision} &= \frac{\mathit{TP}}{\mathit{TP} + \mathit{FP}}, \quad
\mathit{Recall} = \frac{\mathit{TP}}{\mathit{TP} + \mathit{FN}}, \\
\mathit{AP} &= \sum_{n} (R_n - R_{n-1}) P_n.
\end{align}
where $P_n$ and $R_n$ denote precision and recall at the $n$-th operating point. Threshold-dependent performance is reported using accuracy and \textit{F}1-score:
\begin{align}
\mathit{Accuracy} &= \frac{\mathit{TP} + \mathit{TN}}{\mathit{TP} + \mathit{TN} + \mathit{FP} + \mathit{FN}}, \\
F1 &= \frac{2 \cdot \mathit{Precision} \cdot \mathit{Recall}}{\mathit{Precision} + \mathit{Recall}}.
\end{align}
All metrics are computed independently on each test fold, and final results are reported as the mean and standard deviation across folds:
\begin{equation}
\mu = \frac{1}{F} \sum_{f=1}^{F} m_f, \quad
\sigma = \sqrt{\frac{1}{F-1} \sum_{f=1}^{F} (m_f - \mu)^2},
\end{equation}
where $m_f$ denotes the metric value on fold $f$ and $F$ is the total number of folds. As shown in Section~\ref{sec:explainability}, however, accuracy and AUROC alone are insufficient to differentiate leading ALS detection methods, since several baselines converge to statistically indistinguishable performance; this motivates evaluating models along a second, equally critical axis: explainability.

\subsection{Baseline Comparison on EEGET-ALS}
\label{sec:baseline_comparison}
We evaluate Siam-ALS against twelve baseline approaches (A1--A12),
spanning classical ML, DL, and multimodal fusion strategies, under the
identical subject-independent 4-fold cross-validation protocol
(Table~\ref{tab:clinical_comparison}). All baselines reach high AUROC
(0.988--0.998), so differences between them are small. The classical SVM
baselines (A1--A3; AUROC 0.992--0.993) remain competitive with most deep
EEG models: CNN (A4) reaches the highest baseline AUROC (0.998), whereas
LSTM (A5; 0.991) and CNN-LSTM (A6; 0.992) do not exceed the SVMs. Deep
and tree-based models (A4--A10) do, however, achieve somewhat higher
accuracy (0.963--0.974 vs.\ 0.959--0.962). Among the multimodal baselines,
middle fusion (A12), which concatenates EEG and ET embeddings before a
shared classifier, clearly outperforms late fusion (A11), which averages
unimodal probabilities (AUROC 0.998 vs.\ 0.988; accuracy 0.983 vs.\
0.971). This indicates that fusing at the representation level is more
effective than combining decisions. Siam-ALS (A13) matches the highest
AUROC (0.998, tied with A4 and A12) and achieves the best value in every
threshold-dependent metric (accuracy 0.988, MCC 0.976). Fig.~\ref{fig:confusion_matrix}(a)--(b) shows the corresponding confusion
matrices: at window level, errors are small and balanced between false
positives and false negatives, and at subject level, majority voting
correctly identifies all six ALS patients with a single misclassified
control (169/170).

\begin{figure*}[t]
\centering
\includegraphics[width=\columnwidth]{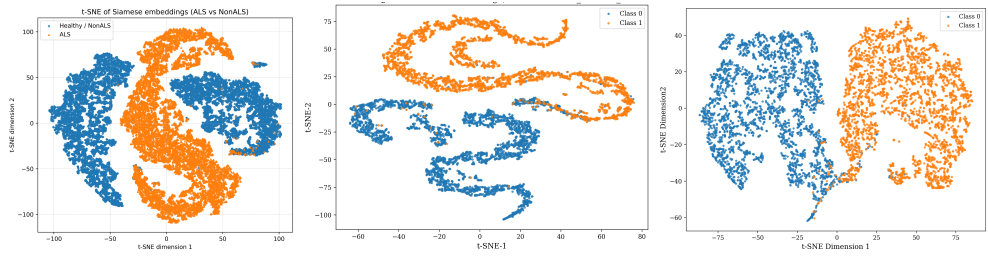}
\caption{
t-SNE visualisation of embedding spaces on the EEGET-ALS dataset.
Left: fused Siam-ALS embeddings learned via metric learning, showing clear separation between ALS and healthy subjects.
 Middle: EEGNet+ET embeddings with late fusion, where EEG and ET features are combined after separate feature extraction.
Right: EEGNet+ET embeddings with mid fusion, where ET features are integrated at an intermediate network layer
}
\label{fig:tsne}
\end{figure*}

Fig.~\ref{fig:tsne} visualizes this qualitatively via t-SNE projections of the learned embedding spaces: Siam-ALS's metric-learned fused embedding shows markedly tighter, better-separated ALS/control clusters than either the late-fusion or middle-fusion embeddings, consistent with the pairwise contrastive objective (Eq.~\eqref{eq:siam3}) shaping a more discriminative representation than direct end-to-end classification training alone. This result is further examined via leave-one-subject-out validation
(Section~\ref{sec:loso}, Table~\ref{tab:loso}), which shows high, though
lower and more variable, per-patient performance for five of the six ALS
subjects, suggesting the 4-fold estimate is not driven by a single
favorable train/test partition.

\begin{figure}[t]
\centering
\includegraphics[width=\columnwidth]{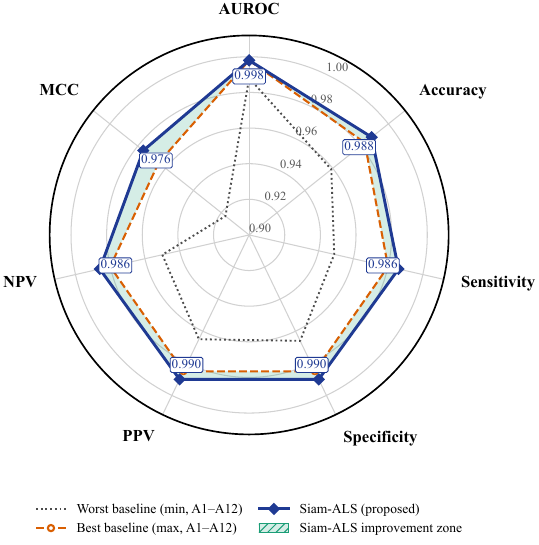}
\caption{Clinical performance envelope of Siam-ALS against the full range of twelve baselines (A1--A12) on EEGET-ALS, across seven clinically relevant metrics (Table~\ref{tab:clinical_comparison}). The dotted inner polygon shows the weakest baseline per metric, the dashed polygon shows the strongest baseline per metric, and the solid indigo polygon shows Siam-ALS. The hatched green region marks where Siam-ALS exceeds the best-performing baseline; on AUROC, Siam-ALS ties the best baselines (A4, A12)}
\label{fig:radar_clinical}
\end{figure}

Beyond AUROC and accuracy, Table~\ref{tab:clinical_comparison} reports
sensitivity, specificity, PPV, NPV, MCC, and each model's explainability
status. Siam-ALS achieves the best or joint-best value in every metric
column and is the only model marked explainable.
Fig.~\ref{fig:radar_clinical} visualizes this across all seven metrics:
the Siam-ALS polygon matches the best baseline on AUROC and exceeds it on
the remaining six axes.

\begin{table*}[t]
\centering
\caption{Clinical performance comparison on EEGET-ALS across all
twelve baselines (A1--A12) and the proposed Siam-ALS (A13), reported
as mean over 4 subject-level folds. Feature indicates the
representation type each model operates on. PPV and NPV are computed on
class-balanced test partitions (50\% ALS prevalence) and therefore do not
reflect clinical prevalence. Sig. reports
uncorrected ($p<0.05$) significance of AUROC
difference from Siam-ALS (Table~\ref{tab:pvals_auc_ttest}) for the
seven directly-tested baselines: \cmark\ = significantly lower AUROC
than Siam-ALS; -- = not significant or not tested; Reference =
comparison baseline. Explainable indicates whether the model
supports attribution over a learned, disease-specific representation}
\label{tab:clinical_comparison}
\setlength{\tabcolsep}{3pt}
\renewcommand{\arraystretch}{1.6}
\scriptsize
\resizebox{\textwidth}{!}{%
\begin{tabular}{c l l c c c c c c c c c}
\hline
\textbf{ID} & \textbf{Model Name} & \textbf{Feature} & \textbf{Explainable} & \textbf{\textit{AUROC}} & \textbf{\textit{ACC.}} & \textbf{Sens.} & \textbf{Spec.} & \textbf{PPV} & \textbf{NPV} & \textbf{MCC} & \textbf{Sig.} \\
\hline
A1  & SVM--PSD                & Hand-crafted EEG & \xmark & 0.992 & 0.959 & 0.951 & 0.966 & 0.965 & 0.952 & 0.918 & \cmark \\
A2  & SVM--PSD + ET           & Hand-crafted EEG & \xmark & 0.993 & 0.962 & 0.958 & 0.967 & 0.967 & 0.958 & 0.925 & \cmark \\
A3  & SVM--Ihar               & Hand-crafted EEG & \xmark & 0.992 & 0.959 & 0.949 & 0.968 & 0.967 & 0.950 & 0.917 & \cmark \\
A4  & CNN                     & Raw EEG          & \xmark & 0.998 & 0.974 & 0.971 & 0.977 & 0.977 & 0.971 & 0.947 & -- \\
\hline
A5  & LSTM                    & Raw EEG          & \xmark & 0.991 & 0.963 & 0.955 & 0.970 & 0.970 & 0.956 & 0.926 & -- \\
A6  & CNN-LSTM                & Raw EEG          & \xmark & 0.992 & 0.968 & 0.963 & 0.972 & 0.972 & 0.963 & 0.936 & -- \\
A7  & Random Forest           & Raw EEG          & Partial & 0.994 & 0.971 & 0.966 & 0.976 & 0.976 & 0.966 & 0.942 & -- \\
A8  & XGBoost                 & Raw EEG          & Partial & 0.993 & 0.970 & 0.964 & 0.975 & 0.975 & 0.964 & 0.939 & -- \\
A9  & Mamba (SSM)             & Raw EEG          & \xmark & 0.993 & 0.973 & 0.968 & 0.977 & 0.977 & 0.968 & 0.945 & -- \\
\hline
A10 & EEGNet                  & Raw EEG          & \xmark & 0.994 & 0.969 & 0.963 & 0.974 & 0.974 & 0.963 & 0.937 & \cmark \\
A11 & EEGNet+ET Late Fusion   & Raw EEG + ET     & \xmark & 0.988 & 0.971 & 0.965 & 0.976 & 0.976 & 0.965 & 0.941 & -- \\
A12 & EEGNet+ET Middle Fusion & Raw EEG + ET     & \xmark & 0.998 & 0.983 & 0.980 & 0.985 & 0.985 & 0.980 & 0.965 & -- \\
\textbf{A13} & \textbf{Siam-ALS (Proposed)} & \textbf{Learned EEG + ET} & \textbf{\cmark} & \textbf{0.998} & \textbf{0.988} & \textbf{0.986} & \textbf{0.990} & \textbf{0.990} & \textbf{0.986} & \textbf{0.976} & \textbf{Reference} \\
\hline
\end{tabular}%
}
{\footnotesize $^{*}$Siam-ALS (A13) is re-trained independently for each
comparison table in this paper and appears again as B3 in the modality
ablation (Table~\ref{tab:modality_ablation}) and as M4 in the architecture
ablation (Table~\ref{tab:ablation}); minor numerical differences across
these three reportings reflect run-to-run training variation rather than
a different model.}
\end{table*}

\subsection{Ablation Studies}

\subsubsection{Data Modality Ablation Study}
Table~\ref{tab:modality_ablation} isolates the contribution of each modality by holding the Siam-ALS architecture fixed and varying only the input streams. The EEG-only variant (B1) achieves 94.3\% accuracy (AUROC 0.9843), confirming that cortical signals alone already carry substantial disease-discriminative information. In sharp contrast, the ET-only variant (B2) performs at chance level (accuracy 49.3\%, AUROC 0.4932, $F_1 = 0.1345$), demonstrating that gaze dynamics alone are entirely insufficient for ALS discrimination in this setting. When the two modalities are fused (B3), performance rises to 98.7\% accuracy and AUROC 0.9985, an absolute AUROC gain of +0.0142 over EEG-only and +0.5053 over ET-only. This asymmetry is itself informative: although ET carries no standalone
discriminative signal, adding it to EEG still improves AUROC by 0.014 and
accuracy by 4.4 percentage points, suggesting that the fusion encoder
exploits cross-modal interactions between cortical and oculomotor
dynamics rather than merely defaulting to the stronger EEG stream. This finding directly motivates the joint EEG--ET explainability analysis, since it confirms that the fused embedding, not EEG alone, is the representation that must be explained to fully account for the model's predictions.

\begin{table*}[t]
\centering
\caption{Data modality ablation study for Siam-ALS, reported as mean $\pm$
standard deviation over 4 subject-level folds. $\Delta$ is the
absolute AUROC gain over the weaker unimodal baseline (ET-only). B3
is architecturally identical to A13 (Table~\ref{tab:clinical_comparison})
and M4 (Table~\ref{tab:ablation}); B1 is comparable to M0
(Table~\ref{tab:ablation}); minor numerical differences reflect
independent training runs (Section~\ref{sec:training_config})}
\label{tab:modality_ablation}
\renewcommand{\arraystretch}{1.5}
\setlength{\tabcolsep}{6pt}
\small
\resizebox{\textwidth}{!}{%
\begin{tabular}{c l l c c c c}
\hline
\textbf{ID} & \textbf{Modality Used} & \textbf{Model Variant} & \textbf{\textit{ACC.}} & \textbf{\textit{AUROC}} & \textbf{\textit{$F_1$}} & \textbf{$\Delta$AUROC vs.\ B2} \\
\hline
B1 & EEG only  & Siam-ALS -- EEG    & $0.943 \pm 0.0063$ & $0.9843 \pm 0.0032$ & $0.9432 \pm 0.0062$ & $+0.4911$ \\
B2 & ET only   & Siam-ALS -- ET     & $0.4932 \pm 0.0044$ & $0.4932 \pm 0.0050$ & $0.1345 \pm 0.0708$ & Reference \\
\textbf{B3} & \textbf{EEG + ET} & \textbf{Siam-ALS fusion} & $\mathbf{0.9872 \pm 0.0010}$ & $\mathbf{0.9985 \pm 0.0001}$ & $\mathbf{0.9841 \pm 0.0010}$ & $\mathbf{+0.5053}$ \\
\hline
\end{tabular}%
}
\end{table*}

\subsubsection{Architectural Ablation Study}
Table~\ref{tab:ablation} isolates the contribution of each architectural component by comparing five Siamese variants (M0--M4) under an identical training protocol, progressively extending a base EEG-only Siamese network with an ET encoder, a dedicated fusion encoder, and an embedding-level logistic-regression classifier. The base model (M0, EEG-only, simple MLP head) already achieves a strong AUROC of 0.988, comparable to the EEG-only result in the modality ablation (B1, AUROC 0.9843 in Table~\ref{tab:modality_ablation}); the small ($<$0.01 AUROC) difference between M0 and B1 reflects independent training runs of an architecturally comparable EEG-only configuration (Section~\ref{sec:training_config}), and both confirm that the Siamese metric-learning formulation alone provides a competitive starting point. Adding depth without ET (M1) slightly reduces performance (AUROC 0.983), showing that naively deepening the head does not help and may overfit under limited subject counts. Activating only the ET encoder (M2) collapses performance to near-chance (AUROC 0.502), consistent with the ET-only result in the modality ablation (B2, AUROC 0.4932 in Table~\ref{tab:modality_ablation}); the small ($<$0.01 AUROC) difference between M2 and B2 reflects independent training runs of an architecturally comparable ET-only configuration (Section~\ref{sec:training_config}), and both results agree that gaze dynamics alone carry essentially no discriminative signal for ALS in this setting. Critically, concatenating EEG and ET embeddings without a dedicated fusion encoder (M3) only recovers EEG-only-level performance (AUROC 0.989), whereas the full model (M4), which adds both the fusion encoder and embedding-level classifier, reaches the best result (AUROC 0.998, $F_1 = 0.988$). The gain from M3 to M4 (AUROC $0.989 \rightarrow 0.998$) reflects the
combined effect of the fusion encoder and the embedding-level classifier;
because both components are added together, this ablation cannot
attribute the improvement to either one alone. It does show, however,
that simple concatenation is insufficient to benefit from ET, and that
the full configuration, which also yields the compact 64-dimensional
embedding used in Section~\ref{sec:explainability}, performs best. In other words, the architectural choices that drive predictive performance are the same choices that make Siam-ALS explainable, not two separate design goals.

\begin{table*}[t]
\centering
\caption{Architectural ablation on EEGET-ALS comparing five Siamese
variants (M0--M4), progressively adding an ET encoder, a fusion
encoder, and an embedding-level classifier to a base EEG-only network,
trained under an identical protocol (Section~\ref{sec:four_fold}). M4 is
architecturally identical to A13 (Table~\ref{tab:clinical_comparison})
and B3 (Table~\ref{tab:modality_ablation}); M2 is comparable to B2,
and M0 to B1 (both Table~\ref{tab:modality_ablation}); minor numerical
differences reflect independent training runs
(Section~\ref{sec:training_config})}
\label{tab:ablation}
\renewcommand{\arraystretch}{1.5}
\setlength{\tabcolsep}{6pt}
\small
\begin{tabular}{c l c c}
\hline
\textbf{ID} & \textbf{Configuration} & \textbf{\textit{AUROC}} & \textbf{\textit{$F_1$}} \\
\hline
M0 & EEG-only, simple MLP head                          & 0.988 & 0.956 \\
M1 & EEG-only, deeper head (no ET)                       & 0.983 & 0.947 \\
M2 & ET-only encoder                                     & 0.502 & 0.142 \\
M3 & EEG+ET concatenation (no fusion encoder)            & 0.989 & 0.962 \\
\textbf{M4} & \textbf{EEG+ET + fusion encoder + emb.\ classifier (full)} & \textbf{0.998} & \textbf{0.988} \\
\hline
\end{tabular}
\end{table*}

\subsection{Statistical Significance Testing}
\label{sec:stats}
To determine whether Siam-ALS's AUROC improvement over each baseline is
statistically meaningful rather than attributable to fold-level noise, we
conduct paired $t$-tests comparing per-fold AUROC values between Siam-ALS
and each of the seven baselines for which per-fold predictions were
directly available: SVM--PSD (A1), SVM--PSD+ET (A2), SVM--IHAR (A3), CNN
(A4), EEGNet (A10), EEGNet+ET Late Fusion (A11), and EEGNet+ET Middle
Fusion (A12). For each baseline $b$, the paired difference
$\Delta_f = \mathit{AUROC}^{\text{Siam-ALS}}_f - \mathit{AUROC}^{b}_f$ is
computed across the $F=4$ subject-level folds, and a two-tailed paired
$t$-test is applied under the null hypothesis $H_0: \mathbb{E}[\Delta_f]
= 0$, with $\mathrm{df} = F-1 = 3$. Because a paired $t$-test with only
four folds has limited statistical power and is sensitive to
non-normality in such a small sample, we additionally report the
non-parametric Wilcoxon signed-rank test as a robustness check;
however, we note that with $F=4$ paired observations the Wilcoxon test
cannot achieve $p<0.05$ even for a fully consistent effect ($p_{\min} = 0.125$; $0.25$ for A4, where $n=3$), and is therefore reported for reference rather than as
a primary significance criterion. Table~\ref{tab:pvals_auc_ttest}
reports the resulting $\Delta$AUROC and $p$-values for both tests, along
with Bonferroni-corrected significance at $\alpha = 0.05/7 \approx
0.0071$ to account for multiple comparisons across the seven baselines.
At the uncorrected level ($p<0.05$), four of the seven baselines (A1, A2,
A3, A10) show a significantly lower AUROC than Siam-ALS, whereas A11, CNN
(A4), and EEGNet+ET Middle Fusion (A12) do not. However, none of these
differences remains significant after Bonferroni correction (smallest
$p = 0.0138 > 0.0071$), so the AUROC advantage of Siam-ALS should be
regarded as suggestive rather than conclusive.

\begin{table*}[t]
\centering
\caption{Paired $t$-test and Wilcoxon signed-rank test comparing Siam-ALS
AUROC against seven baselines across $F=4$ subject-level folds
($\mathrm{df}=3$ for the $t$-test). $\Delta$AUROC represents the mean
paired difference (Siam-ALS $-$ baseline). Sig.\ indicates $p<0.05$
(uncorrected); no comparison remains significant after Bonferroni
correction ($\alpha=0.05/7\approx0.0071$). For CNN (A4), one fold had
zero AUROC difference and was excluded by the Wilcoxon procedure,
leaving $n=3$ pairs.}
	\label{tab:pvals_auc_ttest}
	
	\setlength{\tabcolsep}{6pt}
	\renewcommand{\arraystretch}{1.2}
	\footnotesize
	
	\begin{tabular}{lcccc}
		\toprule
		\textbf{Baseline} & 
		\textbf{$\Delta$AUROC} & 
		\textbf{$t$ ($p$)} & 
		\textbf{Wilcoxon $p$} & 
		\textbf{Sig.} \\
		\midrule
		
		SVM--Ihar (A3)              
		& 0.00658 & 5.21 (0.0138) & 0.125 & \cmark \\
		
		SVM--PSD (A1)               
		& 0.00624 & 4.85 (0.0169) & 0.125 & \cmark \\
		
		SVM--PSD + ET (A2)          
		& 0.00509 & 4.42 (0.0214) & 0.125 & \cmark \\
		
		EEGNet (A10)                
		& 0.00399 & 4.31 (0.0231) & 0.125 & \cmark \\
		
		EEGNet+ET Late Fusion (A11) 
		& 0.01014 & 2.98 (0.0587) & 0.250 & -- \\
		
		CNN (A4)                    
		& 0.00028 & 0.61 (0.5837) & 0.750 & -- \\
		
		EEGNet+ET Middle Fusion (A12)
		& 0.00088 & 0.71 (0.5290) & 0.625 & -- \\
		
		\midrule
		
		\textbf{Siam-ALS (A13)}
		& \multicolumn{3}{c}{\textit{Reference}} &  --\\
		
		\bottomrule
	\end{tabular}
	
\end{table*}

\subsection{Cross-Dataset Generalization on BigP3BCI dataset}
\label{sec:crossdataset}
To assess whether the learned representation generalizes beyond a single
acquisition setup, we evaluate EEG-only Siamese variants (G1--G3,
Table~\ref{tab:bigp3_cross}) on the independent BigP3BCI corpus, which
differs from EEGET-ALS in recording hardware, stimulus paradigm (P300
speller vs.\ gaze-controlled speller), and patient population. The
identical EEG encoder architecture is used without modification
(32 channels, matching the EEGET-ALS input); because BigP3BCI windows
(124 samples at 124~Hz) are four samples shorter than the 128-sample input
the encoder expects, each window is zero-padded to 128 samples prior to
encoding. Trained from scratch on BigP3BCI, the base Siamese model (G1)
reaches an AUROC of 0.954 and a window-level accuracy of 0.890, showing
that the architecture remains effective on a second dataset with a
different paradigm and population. Adding the embedding-level
logistic-regression classifier (G2) raises AUROC to 0.983 and accuracy to
0.937, with sensitivity and specificity improving from 0.882/0.897 to
0.931/0.942. A deeper projection head (G3) yields only a marginal further
gain (AUROC 0.984, accuracy 0.940, sensitivity 0.935, specificity 0.945).
At the subject level, majority voting over each subject's windows improves
on window-level performance for G3, reaching a sensitivity of 0.958 and a
specificity of 0.979. Fig.~\ref{fig:confusion_matrix}(c)--(d) shows the
corresponding confusion matrices: window-level errors are roughly balanced
between classes, with slightly more false negatives than false positives,
and 46 of 48 ALS and 47 of 48 non-ALS subjects are correctly classified.
The gap between AUROC (0.984) and accuracy (0.940) indicates that the fixed
0.5 decision threshold is not optimal on this dataset, so threshold-free
AUROC is the more reliable summary of cross-dataset discrimination.
Because the EEGET-ALS ablation adds the classifier jointly with the fusion
encoder (M3 $\rightarrow$ M4), the G1 $\rightarrow$ G2 comparison is the
only direct evidence in this study that embedding-level classification
improves on the Siamese head alone. The effect of head depth is not
consistent across datasets: it slightly reduced performance on EEGET-ALS
(M0 $\rightarrow$ M1) and slightly improved it here. Since BigP3BCI is
EEG-only, this experiment supports the EEG branch of the architecture but
does not test the ET or fusion components.

\begin{table*}[t]
\centering
\caption{Cross-dataset generalization results on the independent BigP3BCI
corpus using EEG-only Siamese variants (G1--G3), evaluated with the
identical EEG encoder architecture used on EEGET-ALS, without modification.
Window-level metrics are means over 4 subject-level folds; subject-level
metrics use majority voting over each subject's windows (48 ALS / 48
non-ALS). G3 corresponds to Fig.~\ref{fig:confusion_matrix}(c)--(d)}
\label{tab:bigp3_cross}
\setlength{\tabcolsep}{5pt}
\renewcommand{\arraystretch}{1.4}
\small
\resizebox{\textwidth}{!}{%
\begin{tabular}{c l c c c c c c}
\hline
 & & \multicolumn{4}{c}{\textbf{Window level}} & \multicolumn{2}{c}{\textbf{Subject level}} \\
\textbf{ID} & \textbf{Configuration} & \textbf{AUROC} & \textbf{Sens.} & \textbf{Spec.} & \textbf{Acc.} & \textbf{Sens.} & \textbf{Spec.} \\
\hline
G1 & Base Siamese (EEG-only, no classifier head)      & 0.954 & 0.882 & 0.897 & 0.890 & 0.896 & 0.917 \\
G2 & + embedding-level logistic-regression classifier & 0.983 & 0.931 & 0.942 & 0.937 & 0.938 & 0.979 \\
\textbf{G3} & \textbf{+ deeper fusion/projection head} & \textbf{0.984} & \textbf{0.935} & \textbf{0.945} & \textbf{0.940} & \textbf{0.958} & \textbf{0.979} \\
\hline
\end{tabular}%
}
\end{table*}

\begin{figure}[t]
\centering
\includegraphics[width=\columnwidth]{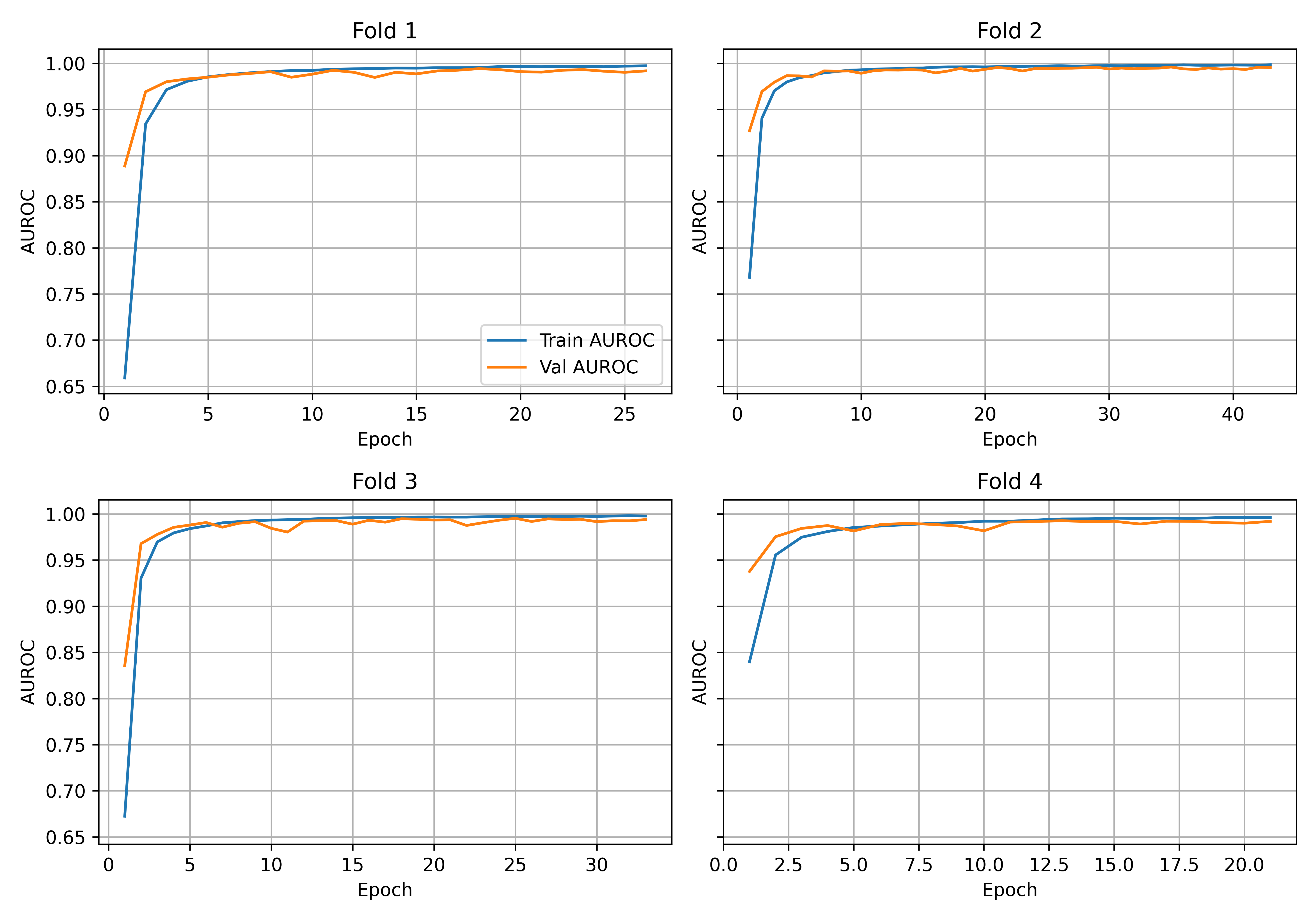}
\caption{Training and validation AUROC across epochs for all four cross-validation folds on EEGET-ALS. The model converges within the first few epochs in every fold and maintains consistently high performance (AUROC $\approx 0.99$), with closely overlapping training and validation curves, indicating stable learning and good generalization without obvious overfitting}
\label{fig:auroc_folds}
\end{figure}

\textit{Model Convergence and Generalization.} Fig.~\ref{fig:auroc_folds} presents the training and validation AUROC trajectories across all four cross-validation folds, confirming that the model converges within the first few epochs and maintains closely overlapping training/validation curves throughout, indicating stable learning without overfitting.

\subsection{Explainability Analysis}
\label{sec:explainability}

As shown in Sections~\ref{sec:baseline_comparison}--\ref{sec:stats}, all twelve baseline models spanning classical ML, DL, and multimodal fusion converge to near-ceiling AUROC (0.988--0.998), and none of the seven directly tested baselines differs significantly
from Siam-ALS after multiple-comparison correction, with three of them
(EEGNet+ET Late Fusion, CNN, and EEGNet+ET Middle Fusion) not significant
even at the uncorrected level (Table~\ref{tab:pvals_auc_ttest}). None of these twelve baselines, however, provide any mechanism to explain why a given sample is classified as ALS. 
This section demonstrates, through six independent attribution methods, including SHAP, LIME, Integrated Gradients, Saliency, Occlusion, and Permutation Importance, applied identically to the frozen embedding-level classifier, two complementary levels of explanation (embedding and signal), and a direct structural comparison against all twelve baselines, that the primary advantage of Siam-ALS is not marginal accuracy improvement but that its architecture is the only one in this comparison that supports rigorous, validated, and clinically verifiable explainability. Table~\ref{tab:attribution_tractability} quantifies this directly, contrasting the exact Shapley attribution cost across representations and showing that only Siam-ALS's compact 64-dimensional embedding admits tractable, exact attribution.

\textit{Benchmarking Explainability Methods.} To identify the most reliable explainability method for the fused embedding space, the results of XAI techniques are compared using fidelity as the evaluation metric. Fidelity was measured by the AUROC retained when a new classifier was trained using only the top 10 ranked embedding dimensions selected by each method from the original 64 dimensional embedding. 

\begin{figure*}[t]
\centering
\includegraphics[width=\columnwidth]{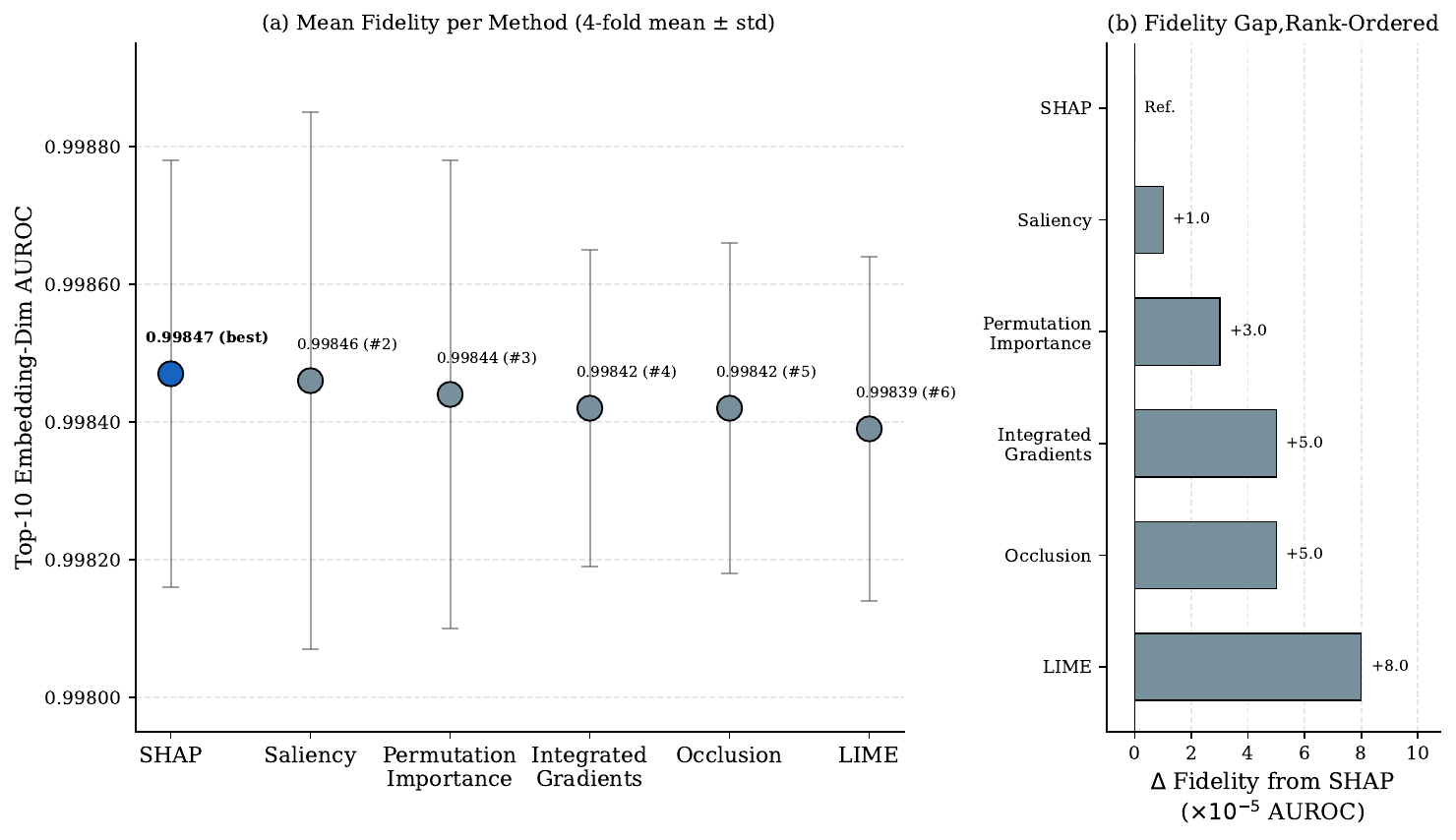}
\caption{Fidelity comparison of six explainability methods applied to the
Siam-ALS fused embedding (Table~\ref{tab:explainability_comparison}).
(a) Mean Top-10 embedding-dimension AUROC per method across 4 folds (error
bars: $\pm$1 SD); SHAP attains the highest mean, but all error bars overlap.
(b) Rank-ordered fidelity gap relative to SHAP, measured as $\Delta$AUROC@10
($\times 10^{-5}$), showing all six methods cluster within a narrow margin}
\label{fig:six_method_fidelity}
\end{figure*}

Table~\ref{tab:explainability_comparison} reports fold-averaged fidelity across all six XAI methods. The XAI comparison showed that all six methods achieved high fidelity, with AUROC values above 0.998, demonstrating that the fused embedding contains stable and interpretable information. SHAP achieved the highest mean fidelity ($0.99847 \pm 0.00031$) using only the top 10 ranked embedding dimensions, although the differences between methods ($\leq 8 \times 10^{-5}$) are far smaller than the fold-level standard deviations. SHAP was therefore at least as faithful as the alternatives and was selected as the primary method for interpreting the EEG- and ET-derived representations learned by Siam-ALS. Fig.~\ref{fig:six_method_fidelity} visualizes this comparison, confirming SHAP's marginal fidelity advantage while showing all six methods achieve near-ceiling, tightly clustered performance.

\begin{tableorg}[t]
\centering
\caption{Fidelity and cross-method agreement of six explainability methods applied to the Siam-ALS fused embedding. Top-10 AUROC reports mean $\pm$ std AUROC retained using only the top-10 ranked embedding dimensions (of 64 total) over 4 folds. Jaccard w/ SHAP reports top-10 dimensional overlap with SHAP, the reference method}
\label{tab:explainability_comparison}
\setlength{\tabcolsep}{5pt}
\renewcommand{\arraystretch}{1.4}
\footnotesize
\resizebox{\columnwidth}{!}{%
\begin{tabular}{l c c}
\hline
\textbf{Method} & \textbf{AUROC@10} & \textbf{SHAP Agreement} \\
\hline
\textbf{SHAP}          & \textbf{0.99847 $\pm$ 0.00031} & \textbf{1.000} \\
Saliency               & 0.99846 $\pm$ 0.00039 & 0.111 \\
Permutation Importance & 0.99844 $\pm$ 0.00034 & 0.538 \\
Integrated Gradients   & 0.99842 $\pm$ 0.00023 & 0.667 \\
Occlusion              & 0.99842 $\pm$ 0.00024 & 0.429 \\
LIME                   & 0.99839 $\pm$ 0.00025 & 1.000 \\
\hline
\end{tabular}%
}
\end{tableorg}

\begin{table*}[t]
\centering
\caption{Attribution tractability underlying the Explainable column
of Table~\ref{tab:clinical_comparison}. Attr. Dim is the
dimensionality attribution must operate over; Exact Shapley Cost is
the asymptotic cost of closed-form Shapley attribution ($\mathcal{O}(2^d)$
in general, linear only for Siam-ALS's compact embedding); Top-10
Fidelity is the AUROC retained using only the top-10 SHAP-ranked
dimensions, reported only where attribution is exact and
physiologically meaningful (Section~\ref{sec:explainability})}
\label{tab:attribution_tractability}
\setlength{\tabcolsep}{4pt}
\renewcommand{\arraystretch}{1.5}
\footnotesize
\begin{tabular}{l c c c}
\hline
\textbf{Representation} & \textbf{Attr. Dim} & \textbf{Exact Shapley Cost} & \textbf{Top-10 Fidelity} \\
\hline
Raw EEG window (A4--A6, A9--A12)       & 4{,}096 & $\mathcal{O}(2^{4096})$ (infeasible) & N/A \\
Hand-crafted PSD/IHAR (A1--A3)          & 16 / 160--169 & $\mathcal{O}(2^{16})$ / $\mathcal{O}(2^{160-169})$\textsuperscript{*} & N/A \\
Tree-based feature importance (A7, A8)  & 4{,}096 & $\mathcal{O}(2^{4096})$ (infeasible) & N/A \\
\textbf{Siam-ALS fused embedding (A13)} & \textbf{64} & \textbf{$\mathcal{O}(d)$ (exact, linear)} & \textbf{0.99847} \\
\hline
\multicolumn{4}{l}{\footnotesize \textsuperscript{*}Exact only for A3 ($d{=}16$); infeasible for A1/A2 ($d{=}160$--$169$). Neither is learned/disease-specific.}
\end{tabular}
\end{table*}

\begin{table*}[t]
\centering
\caption{Performance and explainability comparison across ALS-related and EEG/ET/Siamese methodology precedents. Values reflect each paper's own validation protocol; protocols are listed since results are not directly comparable across rows. All figures are as reported in the original publications and were not independently reproduced or re-evaluated under a common protocol}
\label{tab:comparison_xai}

\renewcommand{\arraystretch}{1.15}
\setlength{\tabcolsep}{2.5pt}

\begin{adjustbox}{max width=\textwidth}
\fontsize{8.5pt}{10pt}\selectfont

\begin{tabular}{c l l l c c l l}
\hline
\textbf{Year} & \textbf{Method} & \textbf{Feature} & \textbf{Dataset} &
\textbf{AUROC} & \textbf{ACC. (\%)} & \textbf{Protocol} &
\textbf{XAI} \\
\hline

2017 & GPFD+Fisher, KNN/LDA~\cite{Liu2017MotorImageryALS}
& GPFD & Private (5 ALS) & -- & 88.3--95.3 & LOO-CV
& Fisher ch.\ sel.\ (partial) \\

2021 & Linear SVC/MLP~\cite{Kocar2021ALS_MRI}
& DTI (FA) + texture & Private ($n{=}502$) & 0.87--0.91 & --
& LOOCV (+holdout for MLP)
& Linear coef.\ (intrinsic) \\

2017 & CNN + RL sampling~\cite{CNNReinforceALS2017}
& CWT/Spectrogram/SPWVD & Public EMG & -- & 96.80
& Train/test split
& None \\

2022 & EEG/ET fusion~\cite{Vortmann2022AttentionHybridBCI}
& Time-series/MTF-GASF images & Private (36 subj.) & --
& 57.2--64.2
& Person-dependent (5x random split)
& None \\

2023 & Siamese AE/CNN~\cite{SiameseSleep2023}
& Raw single-ch.\ EEG & SleepEDF, MASS-SS3 & --
& 85.2/87.2
& 20/31-fold CV
& None \\

2023 & SES+XGBoost/RF~\cite{GeneTargetALS2023}
& Gene expression (RNA-seq)
& Prudencio/Batra RNA-seq & --
& 85.2/88.9
& 9-fold CV
& SHAP \\

2021 & SNN + cosine dist.~\cite{Efe2021SNNSleep}
& Raw single-ch.\ EEG
& Public EEG (Sleep-EDF) & --
& 84.94
& Hold-out
& None \\

2022 & MB-EEGNet+CBAM~\cite{MBEEGCBAM2022}
& Raw EEG (multi-channel)
& BCI-IV2a, HGD & --
& 82.85/95.45
& Within-subj.
& CBAM attn.\ + t-SNE \\

2024 & pcEEG+ (PCA+blink)~\cite{SyncBlinkFusion2024}
& PCA-EEG components + synced blink
& Private (locomotor; loco.+driving)
& --
& 69--78
& 5-rep.\ 80/20
& None \\

2025 & DE+VAE+ET+CNN~\cite{Le2025Attention}
& DE + VAE latent + ET (fix./sacc.)
& Public EEG+ET, healthy subset (n=30)
& --
& 92.4/96.8
& k-fold/cross-subj.
& None \\

2025 & Optimized XGBoost~\cite{OptimDL_ALS2025}
& Clinical/genetic (PRO-ACT vars.)
& PRO-ACT
& 0.955
& 97.96
& 10-fold CV / held-out
& SHAP + feat.\ importance \\

2025 & Pseudo-Siamese~\cite{SSCSleepNet2025}
& Raw single-ch.\ EEG
& 4 public sleep datasets
& --
& MF1 84.5/89.6/89.5/85.4
& k-fold (k=20)
& UMAP \\

2026 & Gated MoE~\cite{ALSGate2026}
& Raw EMG + spectrogram + handcrafted feats.
& EMGLAB (N2001)
& 0.9992
& 99.05
& Subj.-disjoint
& Gating weights \\

\hline

2024 & SVM/CNN/CNN-LSTM~\cite{Ngo2024EEGETALS}
& PSD/IHAR/Raw EEG
& EEGET-ALS (healthy only)
& --
& 88.2--94.5
& Train/test split
& None \\

2024 & SVM/EEGNet/EEG-ITNet~\cite{Ngo2024EEGETALS}
& CSP-BP / EA-processed EEG
& EEGET-ALS (healthy)
& --
& --
& Subj.-held-out
& None \\

\textbf{2026} & \textbf{Siam-ALS (Ours)}
& \textbf{EEG, ET}
& \textbf{EEGET-ALS}
& \textbf{0.998}
& \textbf{98.8}
& \textbf{4-fold subj.-level}
& \textbf{SHAP (linear) + saliency} \\

\hline
\end{tabular}

\end{adjustbox}

\end{table*}

\begin{table*}[t]
	\centering
	\caption{Explainability scorecard for the seven baselines with paired
significance testing against Siam-ALS (Table~\ref{tab:pvals_auc_ttest}).
$\Delta$AUROC and $p$-value are taken directly from
Table~\ref{tab:pvals_auc_ttest}; Explainable is obtained from
Table~\ref{tab:clinical_comparison}. Models are grouped by uncorrected
significance ($p<0.05$); no comparison survives Bonferroni correction
($\alpha=0.0071$). Across both tiers, explainability rather than accuracy
alone is the distinguishing capability of Siam-ALS}
	\label{tab:explainability_scorecard}
	
	\setlength{\tabcolsep}{8pt}
	\renewcommand{\arraystretch}{1.3}
	\footnotesize
	
	\begin{tabular}{lcccc}
		\toprule
		\textbf{Model} & 
		\textbf{$\Delta$AUROC} & 
		\textbf{$p$-value} & 
		\textbf{Sig.} & 
		\textbf{Explainable} \\
		\midrule
		
		\multicolumn{5}{l}{\textit{Tier 1: Lower AUROC than Siam-ALS (uncorrected $p<0.05$)}} \\
		\midrule
		
		SVM--Ihar (A3)               
		& 0.006580 & 0.0138 & \cmark & \xmark \\
		
		SVM--PSD (A1)                
		& 0.006235 & 0.0169 & \cmark & \xmark \\
		
		SVM--PSD + ET (A2)            
		& 0.005087 & 0.0214 & \cmark & \xmark \\
		
		EEGNet (A10)                 
		& 0.003988 & 0.0231 & \cmark & \xmark \\
		
		\midrule
		
		\multicolumn{5}{l}{\textit{Tier 2: No significant difference (uncorrected $p \geq 0.05$)}} \\
		
		\midrule
		
		EEGNet+ET Late Fusion (A11)   
		& 0.010142 & 0.0587 & -- & \xmark \\
		
		CNN (A4)                     
		& 0.000278 & 0.5837 & -- & \xmark \\
		
		EEGNet+ET Middle Fusion (A12)
		& 0.000881 & 0.5290 & -- & \xmark \\
		
		\midrule
		
		\textbf{Siam-ALS (A13)}
		& \multicolumn{3}{c}{\textit{Reference}} & \textbf{\cmark} \\
		
		\bottomrule
	\end{tabular}
	
\end{table*}

Table~\ref{tab:comparison_xai} presents a comprehensive comparison of ALS-related and EEG/ET/Siamese methods in terms of predictive performance, validation protocol, and explainability. Siam-ALS achieves an AUROC of 0.998 and an accuracy of 98.8\% for ALS
detection on EEGET-ALS. Prior work on this dataset, SVM/CNN/CNN-LSTM
models by Ngo et al.~\cite{Ngo2024EEGETALS} (88.2--94.5\% accuracy),
and on healthy-subject EEG--ET data, DE+VAE+ET+CNN~\cite{Le2025Attention}
(92.4--96.8\% accuracy), addressed different classification tasks on
healthy participants rather than ALS versus control discrimination, so
these figures provide context rather than a direct comparison. Some methods report comparable or slightly higher performance, such as Gated MoE~\cite{ALSGate2026}, but these results were obtained using different modalities, datasets, and validation protocols and are therefore not directly comparable. A key distinction is explainability: most prior EEG/ET and Siamese approaches provide no or only partial interpretation, whereas Siam-ALS combines SHAP-based attribution with saliency analysis. Moreover, its Top-10 of 64 embedding dimensions retain an AUROC of 0.998, indicating that most of the discriminative information is concentrated in a compact learned representation.

SHAP is selected as the primary explainability method for two additional reasons: (i) it computes exact Shapley values for the linear embedding-level classifier via \texttt{LinearExplainer}, avoiding the sampling approximations required by LIME or perturbation-based methods, and (ii) it shows perfect top-10 dimensional agreement with LIME (Jaccard $=1.0$), providing cross-method validation from a second, independently-derived attribution technique. Notably, despite comparable fidelity, SHAP and Saliency show low dimensional overlap (Jaccard $=0.11$), indicating that the 64-dimensional fused embedding contains multiple redundant, equally discriminative substructures rather than a single dominant explanatory pathway. This redundancy is itself a desirable robustness property: it means the model's decision is not contingent on any single fragile latent factor, and that multiple independent attribution methods can each recover a sufficient, if not identical, explanatory subset, all of which retain near-ceiling fidelity (Table~\ref{tab:explainability_comparison}).

\begin{figure*}[t]
\centering
\includegraphics[width=\columnwidth]{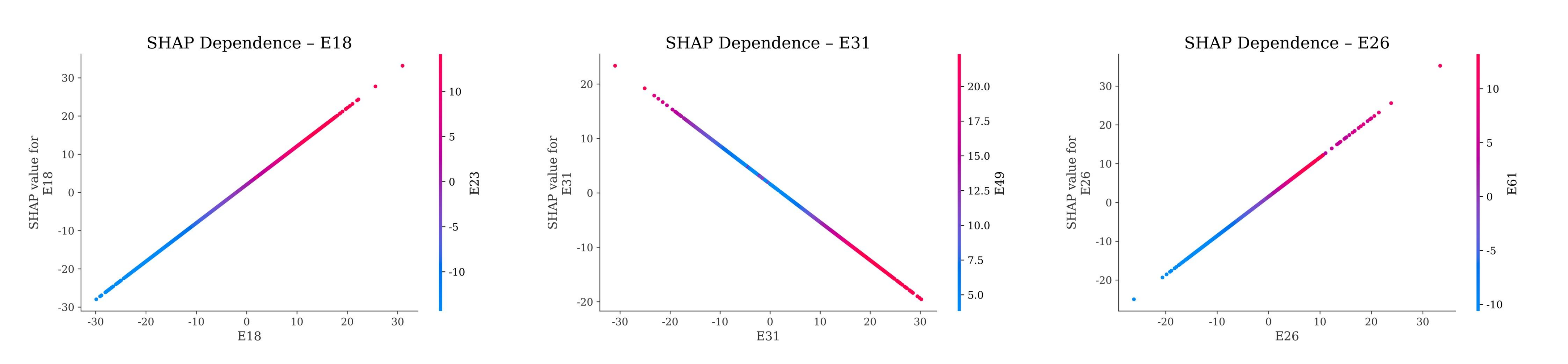}
\caption{SHAP dependence plots for the three top-ranked fused embedding
dimensions (E18, E31, E26) on EEGET-ALS. The y-axis shows each dimension's
SHAP contribution to the classifier output in log-odds. Because the final
classifier is linear, each relation is exactly linear, with slope equal to
the dimension's logistic-regression coefficient: E18 and E26 push predictions
toward ALS, whereas E31 pushes them toward control. Point colour shows the
embedding dimension with the strongest interaction, selected automatically by SHAP}
\label{shap_dependence}
\end{figure*}

\textit{Embedding-Level SHAP Attribution.} Fig.~\ref{shap_dependence} presents SHAP dependence plots for the top-ranked fused embedding dimensions (e.g., E18, E31, E26). These plots reveal a small subset of highly discriminative latent dimensions whose variation shifts the classifier's log-odds linearly and with large magnitude, confirming that the model relies on a compact, interpretable set of learned factors rather than diffusing decision weight uniformly across all 64 dimensions.

\begin{figure*}[t]
\centering
\includegraphics[width=\columnwidth]{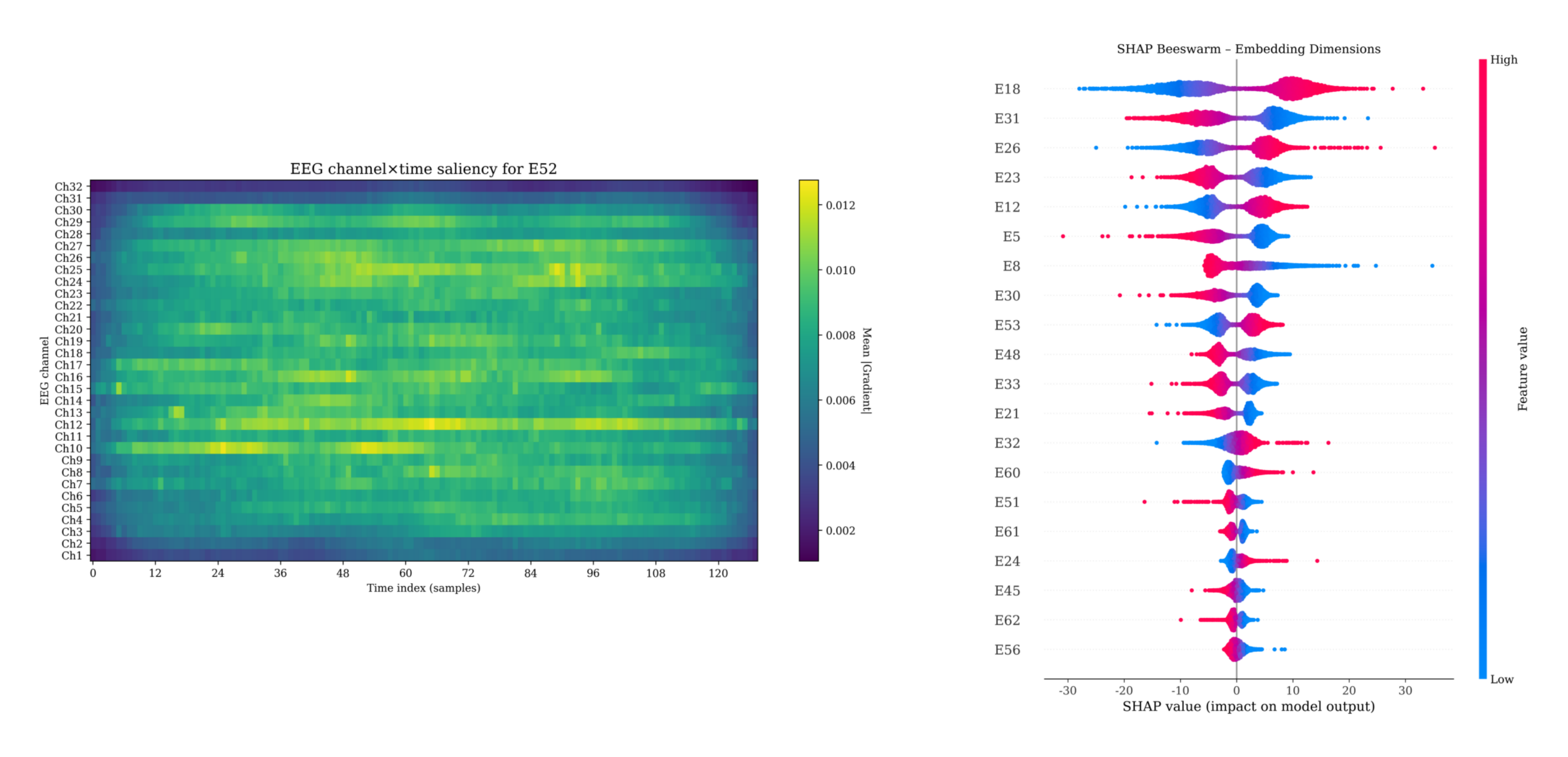}
\caption{Physiological and embedding-level attribution on EEGET-ALS.
(Left) EEG channel$\times$time saliency map (mean absolute gradient) for a
representative fused embedding dimension (E52), illustrating how an
individual embedding dimension draws on specific EEG channels and time
windows. (Right) SHAP beeswarm plot of the 20 highest-ranked fused
embedding dimensions; each point is one window, the x-axis shows its SHAP
contribution to the log-odds output, and colour indicates the dimension's value}
\label{fig:saliency}
\end{figure*}

\textit{Feature-Level Saliency Attribution.} To illustrate how fused
embedding dimensions relate to the underlying physiological signals, we
compute gradient-based saliency maps over the raw EEG input.
Fig.~\ref{fig:saliency} (left) shows a representative example for
embedding dimension E52: saliency is concentrated on a subset of channels
(notably Ch10, Ch12, and Ch24--Ch25) during specific temporal windows
rather than being spread uniformly across the montage, indicating that
individual embedding dimensions draw on spatially and temporally localized
EEG activity rather than diffuse noise. Fig.~\ref{fig:saliency} (right)
complements this with the SHAP beeswarm of the 20 highest-ranked
dimensions, led by E18, E31, and E26.

In summary, Siam-ALS showed a significantly higher AUROC than four of the seven directly tested baselines at the uncorrected level ($p<0.05$), but none of these differences survived Bonferroni correction
(Table~\ref{tab:pvals_auc_ttest}). Since high predictive performance has become comparable among strong models, explainability becomes the key factor that differentiates Siam-ALS from existing approaches. Unlike other baselines, Siam-ALS provides consistent, faithful explanations that localize to specific embedding dimensions and EEG channels through multiple XAI methods (see Table~\ref{tab:clinical_comparison}), supporting its potential as a clinically interpretable ALS detection framework. Table~\ref{tab:explainability_scorecard} consolidates this comparison, showing that across both significance tiers, Siam-ALS is the only model marked explainable.

\section{Conclusion}

This study introduced Siam-ALS, an explainable Siamese DL framework that jointly learns from EEG and ET signals for early ALS detection. By integrating complementary information from EEG-derived neural activity and ET-derived oculomotor behavior, Siam-ALS achieved strong performance on the EEGET-ALS dataset, reaching 98.8\% accuracy and an AUROC of 0.998 under subject-level evaluation. The framework demonstrated competitive performance against a wide range of classical ML and DL models while providing interpretable insights into ALS-related patterns. Beyond accuracy, this study highlights the role of explainability in medical AI. The six XAI methods demonstrated that the learned 64-dimensional embedding captures meaningful ALS-related patterns, with SHAP achieving the highest, though statistically comparable, fidelity in explaining the EEG and ET features that influenced the model’s predictions. Ablation studies showed that multimodal fusion with a dedicated fusion
encoder improves on EEG-only and concatenation variants, while
cross-dataset evaluation on BigP3BCI demonstrated the potential of the
EEG branch to generalize across different BCI settings.

However, this study is limited by the small number of ALS participants in EEGET-ALS, which may affect statistical power and generalization. Future work will focus on larger multi-center datasets and additional BCI paradigms to further validate the framework. Overall, Siam-ALS provides an interpretable and multimodal approach toward reliable AI-based ALS detection.

\section*{Statements and Declarations}

\subsection*{Funding}
This work is partly supported by the Natural Science Foundation of Shanghai, China (Grant No.22ZR1426200 and 17ZR1411900) and the National Natural Science Foundation of China (Grant No. 62103257).

\subsection*{Competing Interests}
The authors declare that they have no competing interests.

\subsection*{Ethics Approval}
This study used publicly available datasets and did not involve any new experiments with human participants. Therefore, ethical approval was not required.
\section*{Data Availability}
The datasets used in this study can be accessed through their respective repositories. The EEGET-ALS dataset, described by Ngo et al.~\cite{Ngo2024EEGETALS}, is available in the Figshare repository at \url{https://doi.org/10.6084/m9.figshare.c.6910027.v1}. The BigP3BCI dataset~\cite{Wong2023bigP3BCI} is publicly available on PhysioNet at \url{https://doi.org/10.13026/0byy-ry86}. The source code will be made publicly available at \url{https://github.com/marahman30104/Siam-ALS} upon acceptance of the manuscript.

\bibliography{main1}

\end{document}